\documentclass[12pt]{article}
\usepackage[letterpaper, left=1in,top=1in,right=1in,bottom=1in]{geometry}
\usepackage[T1]{fontenc} 

\usepackage{amssymb}
\usepackage{pifont}

\usepackage{etoc} 
\usepackage{color}
\usepackage{tikz}
\usetikzlibrary{shapes,decorations,arrows,calc,arrows.meta,fit,positioning}
\tikzset{
    -Latex,auto,node distance =1 cm and 1 cm,semithick,
    state/.style ={ellipse, draw, minimum width = 0.7 cm},
    point/.style = {circle, draw, inner sep=0.04cm,fill,node contents={}},
    bidirected/.style={Latex-Latex,dashed},
    el/.style = {inner sep=2pt, align=left, sloped},
    roundnode/.style={circle, draw=green!60, fill=green!5, very thick, minimum size=7mm},
    squarednode/.style={rectangle, draw=blue!60, fill=blue!5, very thick, minimum size=5mm},
}
\definecolor{red}{HTML}{DB4437}
\definecolor{blue}{HTML}{4285F4}
\definecolor{green}{HTML}{0F9D58}
\definecolor{yellow}{HTML}{F4B400}

\usepackage{makecell}

\usepackage{CJKutf8} 
\usepackage{mathptmx} 

\usepackage{sectsty}
\sectionfont{\centering}

\usepackage[style=apa,backend=biber,isbn=false,uniquename=false,doi=true,url=false,eprint=false,dashed=false,annotation=false]{biblatex}
\AtEveryBibitem{%
  \clearname{translator}%
  \clearlist{language}%
  \clearfield{note}%
}
\bibliography{./LLM_fMRI.bib}

\usepackage{hyperref}
\usepackage{longtable}
\usepackage{multirow}
\usepackage{amsmath}
\mathchardef\mhyphen="2D 

\usepackage{todonotes}
\usepackage{color}
\usepackage{threeparttable}
\usepackage{tabularx}
\usepackage{rotating}
\usepackage{tablefootnote}

\usepackage{siunitx}
\usepackage{titlesec}
\titleformat*{\subsubsection}{\normalsize\itshape}
\usepackage[all]{nowidow}

\usepackage{setspace}
\makeatletter
\newcommand\iraggedright{%
  \let\\\@centercr\@rightskip\@flushglue \rightskip\@rightskip
  \leftskip\z@skip}
\makeatother
\iraggedright

\usepackage[capposition=top]{floatrow}

\usepackage{arydshln} 
\usepackage{appendix}
\usepackage{textcomp}
\usepackage{subcaption}
\usepackage{color,soul}
\usepackage{enumitem} 

\usepackage{ntheorem}
\theoremseparator{:}
\theoremheaderfont{\kern-1cm\normalfont\bfseries}

\usepackage[normalem]{ulem}
\useunder{\uline}{\ul}{}

\usepackage{graphicx}
\graphicspath{{./figs/}}

\usepackage{booktabs}

\newcolumntype{L}[1]{>{\raggedright\let\newline\\\arraybackslash\hspace{0pt}}m{#1}}
\newcolumntype{C}[1]{>{\centering\let\newline\\\arraybackslash\hspace{0pt}}m{#1}}
\newcolumntype{R}[1]{>{\raggedleft\let\newline\\\arraybackslash\hspace{0pt}}m{#1}}

\usepackage{tikz}
\usepackage{pgfplots}
\pgfplotsset{compat=newest}

\usepackage{listings}
\usepackage{xcolor}

\colorlet{punct}{red!60!black}
\definecolor{background}{HTML}{EEEEEE}
\definecolor{delim}{RGB}{20,105,176}
\colorlet{numb}{magenta!60!black}

\def\backtick{\char18}
\lstdefinelanguage{json}{
    basicstyle=\linespread{1}\footnotesize\raggedright\noindent,
    numbers=left,
    numberstyle=\scriptsize,
    stepnumber=1,
    numbersep=8pt,
    showstringspaces=false,
    breaklines=true,
    frame=lines,
    backgroundcolor=\color{background},
    literate=
     *{\{}{{{\color{delim}{\{}}}}{1}
      {\}}{{{\color{delim}{\}}}}}{1}
      {[}{{{\color{delim}{[}}}}{1}
      {]}{{{\color{delim}{]}}}}{1}
      {`}{{{\color{punct}{\backtick}}}}{1},
}

\begin{document}

\thispagestyle{empty}
\newcommand{\mytitle}{Computational Basis of LLM's Decision Making in Social Simulation}

\newcommand{\myabstract}{Large language models (LLMs) increasingly serve as human-like decision-making agents in social science and applied settings. These LLM-agents are typically assigned human-like characters and placed in real-life contexts. However, how these characters and contexts shape an LLM's behavior remains underexplored. This study proposes and tests methods for \emph{probing}, \emph{quantifying}, and \emph{modifying} an LLM's internal representations in a Dictator Game, a classic behavioral experiment on fairness and prosocial behavior. We extract ``vectors of variable variations'' (e.g., ``male'' to ``female'') from the LLM's internal state. Manipulating these vectors during the model's inference can substantially alter how those variables relate to the model's decision-making. This approach offers a principled way to study and regulate how social concepts can be encoded and engineered within transformer-based models, with implications for alignment, debiasing, and designing AI agents for social simulations in both academic and commercial applications, strengthening sociological theory and measurement.}

\begin{titlepage}
    \thispagestyle{empty}

    \begin{center}
        \vspace*{-1cm}

        \singlespacing\textsc{{{\mytitle}}}\\~\\

        {Ji \uppercase{Ma}} \\
        \linespread{1}\small{LBJ School of Public Affairs, University of Texas at Austin\\Gradel Institute of Charity, New College, University of Oxford}\\~\\

        \small{Forthcoming: \textit{Sociological Methodology}; USPTO patent pending.}\\
        \small{First Draft: January 31, 2025; Latest Update: \today}\\

    \end{center}

    \begingroup
    \onehalfspacing

    \begin{abstract}
        \noindent\myabstract
    \end{abstract}


    \endgroup

    \vspace{0.5cm}

    \noindent\rule{4cm}{0.4pt}\\
    \noindent\linespread{1}\footnotesize{\emph{Correspondence}: Ji Ma, 2315 Red River St, Austin, TX 78712, USA; +1-512-232-4240; \href{mailto:maji@austin.utexas.edu}{maji@austin.utexas.edu}. \emph{Acknowledgment}: I thank Chenxin Zhang, Ren\'e Bekkers, and four anonymous reviewers for their constructive comments on earlier versions of the manuscript. \emph{Funding}: The project is partly supported by (1) the Academic Development Funds from the RGK Center, (2) research funds from the Gradel Institute of Charity at the University of Oxford, and computing resources through (3) the Texas Advanced Computing Center at UT Austin \autocite{KeaheyLessonsLearnedChameleon2020} and (4) Dell Technologies, Client Memory Team and AI Initiative PoC Lead Engineer Wente Xiong. \emph{Conflict of Interest}: The author declares no known conflict of interest. \emph{Compliance with Ethical Standards}: The author declares that this study complies with required ethical standards. AI tools were used for grammar, proofreading, and clarity improvement only. \emph{License}: Copyright (c) 2025 Board of Regents, The University of Texas System. The source code and system architecture created by the author are licensed under the CC BY-NC-ND 4.0 license. For commercial use, please contact UT Austin Discovery to Impact \href{mailto:ip@discoveries.utexas.edu}{ip@discoveries.utexas.edu}.}
\end{titlepage}

\begingroup
\onehalfspacing
\normalsize
\etocdepthtag.toc{mtchapter}
\etocsettagdepth{mtchapter}{subsection}
\etocsettagdepth{mtappendix}{none}
\setcounter{tocdepth}{2}

\endgroup

\clearpage
\setcounter{page}{1}
\section{Introduction}

Two aspects of society are particularly interesting to sociologists: structures and meanings. Just as social network analysis supplied formal tools for representing relational structure, computational language models supply new formal methods for studying social meanings \autocite{Arseniev-KoehlerTheoreticalFoundationsLimits2024}. More recently, the Large Language Models (LLMs; e.g., GPT5, Llama, DeepSeek) extend earlier computational text methods, such as dictionaries, topic models, and sentiment classifiers \autocites{EdelmannComputationalSocialScience2020}{NelsonComputationalGroundedTheory2020}, by producing context‑sensitive representations and human-like outputs rather than only aggregate labels \autocite{ZiemsCanLargeLanguage2024,BailCanGenerativeAI2024}. This shift has broadened their role: beyond assistance in representation, coding, or annotation, LLMs are now used as exploratory aides in hypothesis generation and, increasingly, as synthetic respondents in social simulations \autocites{BankerMachineassistedSocialPsychology2024}{AnthisLLMSocialSimulations2025}.

However, the growing use of LLMs in sociology has raised critical questions about their validity, reliability, and theoretical grounding \autocite{KozlowskiSimulatingSubjectsPromise2025,WangLargeLanguageModels2025}. The internal mechanisms driving their decisions remain opaque, and subtle variations in prompting can lead to inconsistent outputs, complicating their use as proxies for human cognition \autocites{AherUsingLargeLanguage2023}{MaCanMachinesThink2024}. This ``black box'' problem is particularly acute for sociologists, who require transparent and theoretically grounded methods to ensure that computational tools genuinely advance our understanding of social phenomena rather than merely reproducing statistical artifacts.

In response, methodological work has largely advanced on two fronts, both of which treat the model as an \emph{opaque system}. The first involves developing sophisticated statistical techniques to correct for model errors. Methods like prediction-powered inference use LLM predictions as auxiliary information to be combined with a small set of ``gold standard'' human data, thereby improving the efficiency of causal estimates \autocite{BroskaMixedSubjectsDesign2025}. Similarly, design-based supervised learning corrects measurement errors in AI-assisted data labeling by leveraging a small number of expert annotations \autocite{RisterPortinariMarancaCorrectingMeasurementErrors2025}. The second approach, prompt engineering, focuses on refining the input (e.g., personas) given to LLMs to elicit more human-like responses \autocite{BisbeeSyntheticReplacementsHuman2024}. While valuable, these methods focus on input-output validation rather than interrogating the internal mechanisms that produce the observed behaviors.

This study introduces a framework for moving inside the black box of LLMs.\footnote{Our method requires access to the model's internal weights and activations, so it is only feasible with open-weight models like Llama and DeepSeek.} Rather than only validating inputs and outputs, we \emph{probe}, \emph{quantify}, and \emph{steer} internal activations tied to sociological concepts \autocite[e.g.,][]{KimLinearRepresentationsPolitical2025}. The central questions we address are: (1) how social meanings are internally represented in LLMs, and (2) how those internal representations can be systematically manipulated to steer model behavior. Drawing on computational sociology, behavioral experiments, and activation engineering, we treat variables (e.g., gender, age, framing) as directions (vectors) in the residual‐stream space of LLMs. We show how to (i) extract these socially meaningful vectors, (ii) orthogonalize them to isolate each variable's unique contribution, and (iii) inject controlled perturbations to modify downstream behavioral decisions in a behavioral experiment. This provides a transparent, theory‑guided workflow for assessing and manipulating how LLMs internally represent social factors, enabling more reliable and interpretable applications in sociological research.~\label{rl:intro_validity} However, this study mainly concerns the \emph{internal representational validity} and controlled steering within the model's activation space. It does not, on its own, establish \emph{external sociological validity} (i.e., accurately representing human cognition and behavior), which requires separate comparative validation with human baselines.

\subsection{LLMs in Sociological Research: Trends and Gaps}

\subsubsection{Measuring ``Meaning'' with Word Embeddings}

Computational text analysis in sociology spans manual content analysis, dictionary methods, and probabilistic topic models \autocite{StoltzMappingTextsComputational2024,NelsonComputationalGroundedTheory2020,CarlsenComputationalGroundedTheory2022}. A pivotal point came with distributional word embeddings, which represent lexical items as points in a high‑dimensional semantic space \autocite{MikolovEfficientEstimationWord2013}. These techniques enabled sociologists to operationalize cultural schemas, semantic fields, and ideologies as measurable geometric relations \autocites{GargWordEmbeddingsQuantify2018}{Arseniev-KoehlerTheoreticalFoundationsLimits2024}{KozlowskiGeometryCultureAnalyzing2019}. Yet classical embeddings suffer from \emph{context invariance}: each word has one static vector regardless of usage, at odds with interactionist and pragmatic traditions emphasizing social and cultural meaning as contextual \autocite[92]{BoutylineMeaningHyperspaceWord2025}.

Transformer-based LLMs (e.g., BERT, GPT variants, Llama) address this by producing \emph{contextualized} token representations across layers. The same surface form now traces different activation trajectories depending on discourse, persona, or pragmatic framing, aligning more closely with sociological accounts of fluid meaning \autocite{MostafaviContextualEmbeddingsSociological2025}. This shift, from static lexical geometry to dynamically evolving internal representations, creates new empirical leverage: researchers can inspect how latent social dimensions (e.g., gendered role expectations, moral frames) are encoded, interact, and transform as text is processed.

Meanwhile, LLM adoption in social science has moved beyond corpus labeling toward end‑to‑end research mediation. Earlier toolkits (topic models, sentiment analyzers, supervised classifiers) uncovered attitudes and sentiments but required labeled data or bespoke training \autocites{GrimmerTextDataPromise2013}{RobertsIntroductionVirtualIssue2016}. Unlike conventional machine learning algorithms, LLMs exhibit remarkable zero-shot learning (performing new tasks without seeing any examples) and few-shot learning (learning effectively from only a handful examples) capabilities \autocite{BrownLanguageModelsAre2020}. Therefore, LLMs can handle new tasks, ranging from text classification to reasoning, without substantial labeled training data. This has eased large-scale text analysis projects in social science, where manually annotated datasets are often limited \autocite{ZiemsCanLargeLanguage2024}. Moreover, researchers have begun to utilize LLMs not just as passive tools for coding and summarizing text but also as active ``collaborators'' in hypothesis generation, literature synthesis, and research ideation \autocites{BankerMachineassistedSocialPsychology2024}{ZhouHypothesisGenerationLarge2024}{BailCanGenerativeAI2024}. In this way, LLMs increasingly shape every stage of empirical inquiry, from initial conceptualization to final reporting \autocite{Chang12BestPractices2024,KorinekAIAgentsEconomic2025}.

\subsubsection{LLMs as Synthetic Respondents: Opportunities and Validity Challenges} \label{sec:llm_opportunities_challenges}

Beyond their role as representational tools, LLMs are increasingly used as ``intelligent agents'' capable of simulating human-like responses in social scenarios. This capability stems from their training on vast datasets of human interaction, which enables them to generate contextually relevant and coherent replies. A growing body of research evaluates LLMs as ``synthetic respondents'' in survey and experimental paradigms \autocites{HortonLargeLanguageModels2023}{ArgyleOutOneMany2023}. In settings like public goods, trust, and dictator games, models have shown prosocial and strategic behaviors that appear consistent with human aggregates \autocites{JohnsonEvidenceBehaviorConsistent2023}{XieCanLargeLanguage2024}{LengLLMAgentsExhibit2024}{MeiTuringTestWhether2024}. Such convergences highlight their potential for rapid, low-cost behavioral prototyping and hypothesis exploration \autocite{AnthisLLMSocialSimulations2025}.

\label{par:challenges}
However, this approach faces significant challenges of validation and external validity. The behavior of LLM agents is often fragile, with subtle shifts in prompts, framing, or situational cues causing drastic changes in their decisions \autocites{AherUsingLargeLanguage2023}{KozlowskiSimulatingSubjectsPromise2025}. Deeper inconsistencies emerge when examining persona manipulations. For example, \textcite{MaCanMachinesThink2024} found that cues for age, gender, and personality elicited behavioral patterns that were inconsistent across different model families and diverged from human distributions. LLMs may generate responses that reflect out-group stereotypes rather than authentic in-group perspectives (i.e., \emph{misportrayal}), and they tend to erase heterogeneity within demographic groups by optimizing for probable, average outputs (i.e., \emph{flattening}) \autocite{WangLargeLanguageModels2025}. From an epistemic perspective, these issues are not merely technical limitations; they risk perpetuating harmful histories of epistemic injustice \autocite{FrickerEpistemicInjusticePower2007}, where the lived experiences of marginalized groups are erased or spoken for by dominant voices \autocite{WangLargeLanguageModels2025,AtariWhichHumans2023}. These discrepancies fuel broader concerns that LLM behaviors may arise from sophisticated statistical pattern completion rather than from internalized causal schemas or a genuine understanding of the world \autocite{LakeBuildingMachinesThat2017}. Consequently, benchmarking model outputs against human aggregates is necessary but insufficient for validation.

Taken together, these challenges caution that apparent behavioral convergence between LLMs and humans may be superficial, arising from text-trained pattern matching rather than human-like cognition. To move beyond validating surface outputs, we distinguish between \emph{internal representational validity}---whether constructs are encoded as meaningful directions that affect a model's internal computations---and \emph{external sociological validity}, which concerns the correspondence of model behavior to human subjects. Our analyses address the former by treating experimental variables (prompt features, persona attributes, instructions) as controlled factors and interrogating how they alter internal representations en route to decisions \autocite{AnthisLLMSocialSimulations2025}. The latter requires dedicated human-LLM comparisons.

\subsubsection{Beyond Prompting: Activation Engineering}

\label{par:prompt_activation}
Prompt engineering is accessible---simple text edits can shift outputs---but it is fragile and entangles style, content, and persona, making effects hard to separate or replicate across paraphrases. When the goal is theory, convenience is not enough: we must test whether observed differences reflect stable internal dimensions rather than surface cues. Activation engineering makes that trade worthwhile by intervening in the residual stream to localize where a factor acts, quantify its alignment with the decision direction, and disentangle variables for causal probing \autocite{TurnerSteeringLanguageModels2024,SubramaniExtractingLatentSteering2022}. It supports \emph{methodological robustness}: computationally skilled sociologists can audit internal validity, estimating, orthogonalizing, and stress‑testing concept vectors, so that the broader field can later deploy prompt‑level workflows with greater confidence and without worrying about the model's internal validity.

Specifically, our activation engineering (i) extracts internal differences associated with theoretically defined contrasts, (ii) orthogonalizes them to isolate unique social dimensions, and (iii) re‑injects controlled perturbations to test causal influence on downstream decisions. This shifts analysis from external prompting to mechanism probing and addresses three sociological needs:

\begin{enumerate}[topsep=0pt,itemsep=0ex,partopsep=0ex,parsep=0ex,label=(\arabic*)]
    \item \emph{Concept operationalization}: Social variables (gender persona, interaction horizon, framing) become measurable vectors whose geometry (angles, projections) encodes relational structure.
    \item \emph{Causal probing}: Injecting purified (orthogonalized) vectors tests whether a latent representation exerts directional pressure on decisions, moving beyond correlational output validation.
    \item \emph{Transparency and auditability}: Layer‑wise trajectories reveal where (depth) and how (magnitude, alignment) social factors enter decision formation, informing bias audits and theoretical interpretation.
\end{enumerate}

In this study we instantiate this framework within a Dictator Game, demonstrating how variables with social meanings can be: (1) extracted from residual streams, (2) disentangled via orthogonalization, and (3) applied as precise steering interventions. This establishes a methodological bridge between social/behavioral sciences and computational modeling, advancing a more interpretable, mechanism‑aware computational social science. While this improves interpretability and control of internal mechanisms, it does not by itself justify inferences about human social behavior without external validation against human baselines.

\subsection{Architecture of LLMs: A Less Technical Overview}

LLMs are designed to predict text token by token, much like a sophisticated version of your smartphone's autocomplete feature. When you see it finish a sentence, it is essentially guessing, based on patterns it has learned, which word (or token) is most likely to come next. For example, imagine you start typing: ``The cat sat on the \dots'' A simple guess might be ``mat,'' since that's a common phrase. But LLMs can go further, understanding that words like ``floor'' or ``sofa'' might also fit depending on the broader context. It ``knows'' this because it has been trained to find patterns in massive amounts of text and then to predict how real sentences typically continue.

We can use the \texttt{Llama3.1-8B} model as an example of how LLMs operate because its architecture and training methodology are representative of the transformer-based language models---the dominant architecture of LLMs \autocite{VaswaniAttentionAllYou2017}. The Llama family of models, originally open-sourced by researchers aiming to democratize access to powerful LLMs, has gained wide adoption in both academic and industry settings.

\begin{sidewaysfigure}[htbp]
    \centering
    \caption{\textsc{Illustration of Llama model internal structure}} \label{fig:llm_internal}
    \includegraphics[width=1\textwidth]{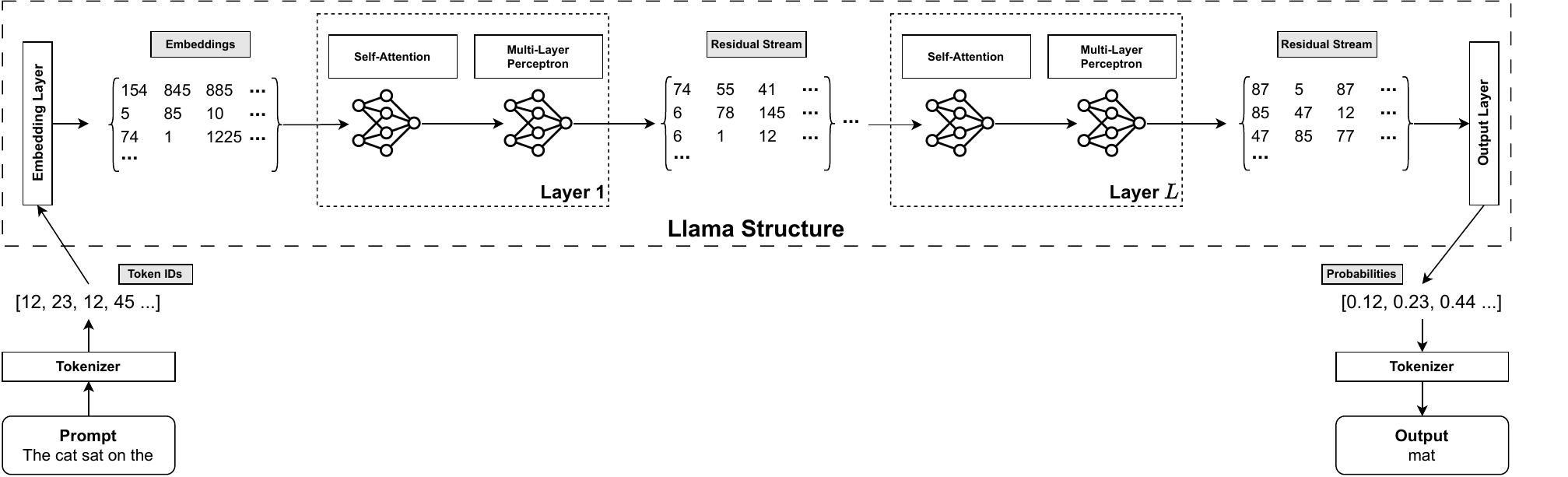}
    \begin{minipage}{1\textwidth}
        \footnotesize
        \textit{Notes}: For \texttt{Llama-3.1-8B-Instruct}, \(L=32\).
    \end{minipage}
\end{sidewaysfigure}

In broad terms, \texttt{Llama3.1-8B} follows a pipeline that begins by converting textual input into numerical form (i.e., token IDs; \ref{sec:tokenizer}), converting these token IDs into high-dimensional embeddings (\ref{sec:embedding_layer}), passing those embeddings through multiple \emph{decoder layers} that refine the representation (\ref{sec:decoder_layers}), and finally using a \emph{head} layer to guess the next most probable token (\ref{sec:output_layer}). Below, we delve deeper into the mechanics of this pipeline. Along with the illustration in Figure \ref{fig:llm_internal}, this explanation demonstrates how the model transforms input text into meaningful predictions.

\subsubsection{Tokenizer: Turning Words into Numeric IDs} \label{sec:tokenizer}

The first step is to convert raw text into discrete units (\emph{tokens}) that the model can process, a procedure called \textit{tokenization}. The tokenizer assigns each token a unique integer ID. For illustration, ``cat'' might map to 3{,}456 and ``mat'' to 7{,}891. Modern LLMs (including \texttt{Llama3.1-8B}) use subword tokenization (a Byte Pair Encoding‑style algorithm) rather than whole‑word vocabularies \autocite{SennrichNeuralMachineTranslation2016}. Rare or morphologically complex words are decomposed into frequent substrings so the model can represent virtually any input without an enormous word list. For example, ``sociologically'' might be segmented into smaller meaningful pieces (e.g., ``socio'', ``logical'', ``ly''), though exact splits depend on the trained tokenizer. The \texttt{Llama3.1-8B} tokenizer has a fixed vocabulary of 128{,}256 learned tokens (including whole words, subwords, symbols, and control tokens), each mapped deterministically to an ID. These IDs are the numerical sequence passed to the embedding layer.

\subsubsection{Embedding Layer: Mapping Tokens IDs to Vectors} \label{sec:embedding_layer}

Once the input text is tokenized, each token ID is mapped to a high-dimensional vector known as an \textit{embedding}. This is where the model begins to represent the semantic meaning of the tokens. In \texttt{Llama3.1-8B}, each token is mapped to a 4,096-dimensional vector. These embeddings are not static; they are learned during the model's training process and are updated as the model processes the text. This ``contextualization'' of embeddings is a key feature of transformer models, distinguishing them from earlier methods like Word2Vec, where each word had a fixed embedding. The initial embeddings are then combined with positional encodings, which provide the model with information about the order of the tokens in the sequence.

\subsubsection{The ``Thinking Blocks'': Stacked Decoder Layers} \label{sec:decoder_layers}

The ``heavy lifting'' in \texttt{Llama3.1-8B} happens in \emph{32 stacked Decoder Layers}. Each layer refines the representation of the input text (i.e., the embedding matrix) and passes the updated matrix, the \emph{residual stream}, to the next layer. Each decoder layer mainly has two critical components: \emph{Self-Attention Mechanism} and \emph{MLP (Multi-Layer Perceptron)}. Below we intuitively introduce the three major components briefly:

\paragraph{Self-Attention Mechanism.}

This component helps the model understand which words or tokens in a sentence should be most relevant to each other. For instance, if a sentence discusses ``cats chasing mice,'' the model can learn to focus on words like ``chase,'' ``feline,'' or ``mouse'' in the right context, even if they appear several words apart. This mechanism allows the model to capture long-range dependencies and understand the relationships between different parts of the text. This design is a key reason why Llama models can generate coherent and contextually relevant sentences and a major breakthrough in natural language processing \autocite{VaswaniAttentionAllYou2017}.

\paragraph{Multi-Layer Perceptron (MLP).}

The output of the self-attention mechanism is then passed to a MLP, also known as a Feed-Forward Network (FFN). The MLP is a relatively simple neural network that applies a series of non-linear transformations to the attention output. Its role is to process the information aggregated by the self-attention mechanism and to add representational capacity to the model. While self-attention is responsible for identifying relationships between tokens, the MLP is where much of the model's ``knowledge'' is stored and processed. It allows the model to learn complex patterns and relationships in the data that go beyond the pairwise comparisons of the attention mechanism.

\paragraph{Residual Streams (or Skip Connections).}

After each sub-block (Self-Attention or MLP), the output is added back into the original input of that sub-block. This creates a ``residual stream,'' also called a ``skip connection.'' It ensures that each layer can refine existing information without discarding what was learned previously. In our ``The cat sat on the ...'' example, a residual connection ensures that early-layer knowledge about ``cat'' and ``mat'' persists as later layers weigh in. This ``accumulated knowledge'' approach makes training deep networks more stable and effective.

\subsubsection{Final Output Layer: Predicting the Next Token.} \label{sec:output_layer}

Once the text data flows through all 32 decoder layers, the final layer, labeled as \emph{lm\_head} in \texttt{Llama3.1-8B}, translates the 4,096-dimensional representation into a probability distribution across 128,256 possible tokens (e.g., \([0.12, 0.23, 0.22 ...]\), a list of 128,256 possibilities,\footnote{These numbers are logits before transformation with, for example, softmax function.} the index position of each possibility corresponds to a unique token). This is where the model ultimately decides what the next word or token should be, given everything it has seen so far. In our example, the model might predict that the next word after ``The cat sat on the ...'' is ``mat'' with a 70\% probability, ``floor'' with a 20\% probability, and so on.

Through above architecture, \texttt{Llama3.1-8B} can generate coherent and contextually relevant sentences, paragraphs, or even long-form responses. Whether it is predicting ``mat'' when you type ``The cat sat on the ...'' or handling more intricate tasks, such as summarizing lengthy documents, answering specialized-domain questions, or crafting creative narratives, the model relies on a pipeline of embeddings, attention, MLP, and residual connections, each contributing to the advanced language understanding and generation system.

\subsubsection{Relation to Other LLMs} \label{sec:relation_to_other_llms}

We use \texttt{Llama3.1-8B} as an illustrative case because its transformer architecture is representative of modern open-weight LLMs \autocite{VaswaniAttentionAllYou2017}. Key architectural elements our method relies on---tokenization, embedding layers, stacked decoders with attention and feedforward sub-blocks, and a residual stream with additive updates---are common across a wide range of models. Our steering procedure operates on the geometry of the residual stream (Sections \ref{sec:decoder_layers} and \ref{sec:additive_nature}), an architectural invariant of decoder-only transformers. Therefore, our control mechanism is, in principle, portable across models.

However, portability does not imply uniform effects. Models vary in layer depth, hidden dimensionality, normalization, gating, and training data. For instance, state-of-the-art models like \texttt{DeepSeek-R1} use mixture-of-experts (MoE) feedforward blocks, where tokens are routed to specialized experts, altering sparsity and local subspace structure \autocites{LepikhinGShardScalingGiant2020}{DaiDeepSeekMoEUltimateExpert2024}{DeepSeek-AIDeepSeekR1IncentivizingReasoning2025}. Diverse pre-training and instruction-tuning also change how and where social concepts are encoded. Our approach can accommodate these differences by re-estimating all steering vectors for each model and depth. In theory, as long as residual updates are approximately additive and the decision subspace is present, injecting the DV-aligned component of a partial IV vector will produce controlled behavioral shifts. In practice, the magnitude, sign stability, and layer sensitivity of these shifts are empirical and may vary with model family and data.

\subsection{Key Methodological Background}
\subsubsection{Representation of Social Concepts in LLMs} \label{sec:linear_representation}

Recent studies have shown that LLMs encode social concepts \autocite{KimLinearRepresentationsPolitical2025,ParkLinearRepresentationHypothesis2024,TiggesLinearRepresentationsSentiment2023} and facts \autocites{EngelsNotAllLanguage2024}{GurneeLanguageModelsRepresent2024} in their internal space, either as one-dimensional (e.g., good vs.\ bad) or multi-dimensional (e.g., seven days of the week). Concretely, the model's internal representation of, say, ``male'' or ``female'' can be viewed as a \emph{vector} in a high-dimensional space. By nudging this vector in a particular direction, one can effectively steer the model's output, for instance, adding a small offset to the ``male'' vector might make the model more likely to adopt male-coded traits or perspectives in its text.

Because these transformations manifest as vectors in the residual‐stream space, we can isolate and control each variable's unique contribution by defining corresponding vectors (see Figure \ref{fig:vector_diagram} and Section \ref{sec:vector_variable_variation}). This means that a vector capturing ``age 20 to age 40,'' ``female to male,'' or ``neutral to liberal'' can be treated with standard vector operations, such as addition, subtraction, and projection, without loss of interpretability. The additive structure in the LLM's hidden layers further enables targeted steering via small, well‐chosen modifications to the residual stream, thereby allowing us to manipulate the model's behaviors (Section \ref{sec:additive_nature}).

\subsubsection{Vector of Variable Variation} \label{sec:vector_variable_variation}

Figure~\ref{fig:vector_diagram} provides an intuitive geometric view of how we treat each variable's influence as a \emph{vector} in the LLM's residual-stream space. In this simplified 2D illustration, moving from \emph{age 20} to \emph{age 40} at layer \(\ell\) appears as a vector in the ``Age'' direction. This vector is obtained by subtracting the average residual stream of layer \(\ell\) for all trials with age 20 from the average residual stream of the the same layer for all trials with age 40. Similarly, subtracting the average residual stream of layer \(\ell\) for all trials with \(D=0\) from the average residual stream of the same layer for all trials with \(D=10\) yields an vector in the ``Decision'' direction. We can then use standard vector operations to quantify interactions among these variables:

\begin{figure}[htbp]
    \centering
    \caption{\textsc{Illustration of Variable Vectors at Layer \(\ell\)}} \label{fig:vector_diagram}
    \includegraphics[width=.8\textwidth]{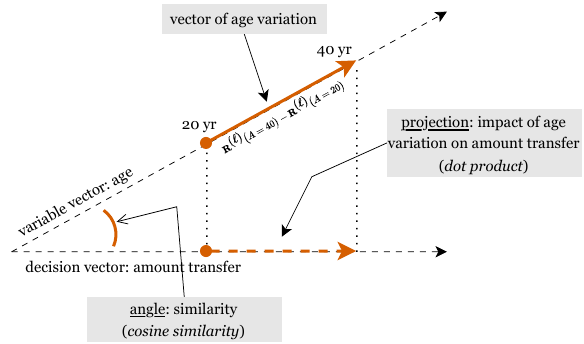}
    \begin{minipage}{.8\textwidth}
        \footnotesize
        \textit{Notes}: \(\mathbf{R}^{(\ell)}\bigl(A=20\bigr)\) is the average residual stream of layer \(\ell\) for all trials with age 20.
    \end{minipage}
\end{figure}

\paragraph{Angle (cosine similarity).}

The cosine similarity between two vectors measures the cosine of the angle between them. It is a measure of directional similarity, ranging from -1 (opposite directions) to 1 (same direction), with 0 indicating orthogonality. We use cosine similarity to assess whether the representations of two concepts are aligned in the model's activation space.

\label{par:cosine_similarity_issue}
However, while cosine similarity is a common way to check whether two concepts ``point'' in a similar direction, it needs to be used with caution. As shown by \textcite{SteckCosineSimilarityEmbeddingsReally2024}, the way embeddings are learned and regularized can stretch or shrink the hidden axes of the space without changing the model's actual predictions. After this stretching, normalizing vectors (as cosine does) can produce arbitrary, and sometimes non‑unique, ``similarities,'' even though the underlying dot products remain well‑defined. In deep models, combinations of regularization methods can implicitly rescale different latent dimensions, making cosine values even harder to interpret.

To address this in our study, we explicitly separate ``direction'' from ``strength.'' We use cosine similarity only as a directional indicator, and we pair it with the dot product, which also reflects magnitude. Looking at both alignment (cosine) and magnitude (dot product) gives a more complete picture of how variables relate inside the model and avoids over‑interpreting cosine in isolation (see our results on the dissociation between the two measures, i.e., Figure \ref{fig:relations_IV_DV_steering_vectors}).

\paragraph{Projection and dot product.}

The dot product of two vectors is related to the cosine similarity but is scaled by the magnitudes of the vectors. Specifically, the dot product of vector \(\mathbf{a}\) and \(\mathbf{b}\) is \(\|\mathbf{a}\|\|\mathbf{b}\|\cos\theta\). We use the dot product to measure the magnitude of the projection of one vector onto another. This allows us to quantify how much a change in one variable (e.g., age) contributes to a change in another (e.g., the decision to give money).

Mathematically, we can define the \emph{vector of variable variation}. Suppose we want the vector representing how a single variable \(X\) shifts the LLM's residual‐stream representation at layer~\(\ell\).  For concreteness, let \(X=\text{age}\) with two distinct values: \(A=20\) and \(A=40\). In each experimental trial, we hold the variable of interest (in this case, age) constant while randomizing all other variables (gender, game instruction, future interaction). This experimental design allows us to isolate the average effect of the variable of interest, as the effects of the other, randomized variables will average out to zero over a large number of trials. This is a standard approach in experimental research and is preferable to holding all other variables constant, as it allows us to estimate the average effect of a variable in the presence of other sources of variation.

We then collect two sets of residual‐stream vectors: (1) \(\bigl\{\mathbf{r}^{(\ell)}_1,\ldots,\mathbf{r}^{(\ell)}_{n_1}\bigr\}\), from all trials where \(A=20\) (but \(G,I,M\) vary randomly); (2) \(\bigl\{\mathbf{r}^{(\ell)}_{n_1+1},\ldots,\mathbf{r}^{(\ell)}_{n_1+n_2}\bigr\}\), from all trials where \(A=40\) (again with \(G,I,M\) randomized).

We compute the mean residual stream for each set:
\[
    \mathbf{R}^{(\ell)}\bigl(A=20\bigr)
    \;=\;
    \frac{1}{n_1}\!\sum_{j=1}^{n_1}\mathbf{r}^{(\ell)}_j,
    \quad
    \mathbf{R}^{(\ell)}\bigl(A=40\bigr)
    \;=\;
    \frac{1}{n_2}\!\sum_{k=n_1+1}^{n_1+n_2}\mathbf{r}^{(\ell)}_k.
\]

By taking the \emph{difference} of these means, we obtain the \emph{vector of age variation}:
\[
    \mathbf{v}_\text{age}
    \;=\;
    \mathbf{R}^{(\ell)}\bigl(A=40\bigr)
    \;-\;
    \mathbf{R}^{(\ell)}\bigl(A=20\bigr).
\]

Because the other variables are randomized and we have a large number of experimental trials, \(\mathbf{v}_\text{age}\) reflects \emph{how shifting from age~20 to age~40 changes the internal representation on average}, after ``averaging out'' any incidental effects of gender, game instruction, or future interaction on residual streams. In the same way, we can define analogous vectors for gender (male vs.\ female), game instruction (give vs.\ take), or future interaction (meet vs.\ not meet) by grouping trials according to whichever variable is under study and then subtracting the mean residual streams. Although Figure~\ref{fig:vector_diagram} shows only a two-dimensional plane for clarity, the same vector operations extend naturally into the higher-dimensional spaces of transformer-based LLMs.

Our method builds directly on the concept of ``steering vectors'' from the activation engineering literature \autocite{TurnerSteeringLanguageModels2024,KimLinearRepresentationsPolitical2025}. However, our work makes a key sociological contribution. While most existing work focuses on steering broad, often abstract concepts (e.g., ``romance,'' ``violence''), we develop a methodology for extracting and manipulating steering vectors that correspond to \emph{specific and socially meaningful variables} within a \emph{controlled experimental design}. This allows us to move from a general ability to steer LLMs to a specific, sociologically informed method for testing hypotheses about the influence of social factors on behavior. This is the core innovation of our approach: we are not just steering LLMs, we are conducting virtual social experiments at variable level inside them.

\subsubsection{Additive Nature of Internal Representations in LLMs} \label{sec:additive_nature}

Modern transformer-based LLMs maintain a \textit{residual stream} \(\mathbf{r}^{(\ell)}\) at each layer \(\ell\). These residual streams accumulate the hidden representation of the input prompt as it propagates through attention and feedforward sub-layers. Crucially, in many transformer designs, these updates occur in a largely \textit{additive} fashion:
\[
    \mathbf{r}^{(\ell+1)}
    \;\approx\;
    \mathbf{r}^{(\ell)}
    \;+\;
    \text{(some transformation of \(\mathbf{r}^{(\ell)}\))}.
\]

Because of this structure, small, well-chosen interventions in \(\mathbf{r}^{(\ell)}\) can exert a relatively \emph{predictable} effect on the final output (the LLM's generated text or decision). Specifically, by carefully inserting a \emph{steering vector} \(\Delta\) (see Section \ref{sec:iv_steering_vectors}) into \(\mathbf{r}^{(\ell)}\), we nudge the hidden representation in a direction and magnitude of interest. Subsequent layers preserve much of this directional influence due to the additive nature of the residual streams.

To give a concrete example, imagine an LLM-based Dictator that must decide how much money to give (or take) to another player. For a simplified two-layer transformer, the first layer produces \(\mathbf{r}^{(1)}\), which is then mostly additively updated in the second layer to yield \(\mathbf{r}^{(2)}\). If we add a small steering vector \(\Delta\) to \(\mathbf{r}^{(1)}\) at layer~1---perhaps nudging the model toward generosity---\(\Delta\) ``persists'' onto layer~2 through the residual connection and ultimately shifts the final decision. In this way, targeted additions to the residual streams can directly manipulate the model's internal reasoning trajectory and shape its final decision.

\subsection{Overview of Experiment Design}

We systematically \textit{probe} (Step 1), \textit{quantify} (Steps 2--4), and \textit{modify} (Step 5) how LLM-based agents behave in a Dictator Game (\ref{sec:baseline_experiment}). This approach provides not only an \textit{analytical} framework (decomposing the influence of different factors on final decisions) but also a \textit{practical} one (actively steering the model's choices). Overall, our approach consists of the following steps:

\begin{enumerate}[topsep=0pt,itemsep=0ex,partopsep=0ex,parsep=0ex,label=(\arabic*)]
    \item Focus on Residual Streams (\ref{sec:decoder_layers}): We first record the \textit{residual streams} because their additive nature (\ref{sec:additive_nature}) allows small ``nudges'' to predictably shift the model's internal representations (\ref{sec:vector_variable_variation}).
    \item Identify Steering Vectors (\ref{sec:basic_steering_vector} and \ref{sec:dv_steering_vector}): For each factor \(X\), we measure how the LLM's internal representation changes when toggling \(X\) (e.g., female vs. male).
    \item Partial Out Confounds (\ref{sec:partial_steering_vector}): By subtracting overlapping components, we obtain a \textit{pure} vector capturing only \(X\)'s unique effect (uncontaminated by age, framing of games, etc.).
    \item Project onto Decision Space(\ref{sec:projection_partial_iv_steering}): We project the partial vector of \(X\) onto LLM's decision vector, so that we can measure the impact of each factor on the LLM's final decision.
    \item Manipulate (\ref{sec:precise_control_ivs}): We inject a scaled version of the projection into the residual streams to modify \(X\)'s impact on LLM's final decision, allowing us to steer the model's behavior.
\end{enumerate}

In summary, by \textit{quantifying} steering vectors, \textit{orthogonalizing} them for pure effects, and then \textit{injecting} the scaled projections back into the LLM's residual streams, we can manipulate an LLM agent's decision in the Dictator Game.

\section{Methods}

\subsection{Baseline Experiment} \label{sec:baseline_experiment}

\subsubsection{Game Setup}

We followed the game setup described in \textcite[12-13]{MaCanMachinesThink2024} and devised a simplified version with fewer independent variables to better focus on the main tasks of this study (i.e., \textit{probe}, \textit{quantify}, and \textit{modify}). In this Dictator Game, an LLM-based ``dictator'' decides how much to allocate (or ``take'') from another player (the ``recipient''). Both the initial endowment and the amount that can be transferred are set to \$20. For example, if the dictator transfers \$10 to the recipient, both players will end up with \$30, which is considered a fair split.

Below are the independent variables considered in the baseline experiment (i.e., for probing and quantifying purposes, no manipulation):

\begin{enumerate}[topsep=0pt,itemsep=0ex,partopsep=0ex,parsep=0ex,label=(\arabic*)]

    \item Agent Persona: Gender \(G \in \{\text{male}, \text{female}\}\) and Age \(A \in \{20, 21, 22, \dots, 60\}\).
    \item Game Instruction \(I \in \{\text{give}, \text{take}\}\): In the \textit{give} condition, the dictator can only transfer a nonnegative amount \(D \in [0, 20]\). In the \textit{take} condition, the dictator can transfer a negative amount (effectively ``take'' from the recipient), so \(D \in [-20, 20]\).
    \item Future Interaction \(M \in \{\text{meet}, \text{not meet}\}\): Whether the recipient is a stranger the agent will meet afterward (``stranger\_meet'') or not (``stranger'').

\end{enumerate}

Let the dependent variable (DV) be the amount \(D\) that the LLM-based dictator ultimately give (positive) or takes (negative). The model's input includes \((G, A, I, M)\) plus any additional invariant prompt context. Formally,

\[
    D \in
    \begin{cases}
        [0, 20],   & \text{if } I = \text{give}, \\
        [-20, 20], & \text{if } I = \text{take}.
    \end{cases}
\]

\subsubsection{Baseline Trials: Randomizing All Variables}

In each baseline trial, we (1) randomize the values of all the input variables: \(G\) (gender), \(A\) (age from 20 to 60), \(I\) (give vs. take), and \(M\) (meet vs. not meet), and (2) collect the LLM's responses (i.e., the amount transferred \(D\)). We did not modify any of the model's generation hyperparameters, such as temperature or top-p. This yields an empirical distribution:
\[
    p(D \,\mid\, G, A, I, M).
\]

To collect sufficiently robust data:

\begin{enumerate}[topsep=0pt,itemsep=0ex,partopsep=0ex,parsep=0ex,label=(\arabic*)]
    \item We obtain a set of responses (i.e., trials) \(D_1, D_2, \dots, D_k\), where \(k = 1,000\).
    \item The LLM can be prompted multiple times with the same tuple \((g, a, i, m)\) because the total number of input combinations (i.e., 320) is much smaller than the total number of trials (i.e., 1,000).
    \item We can compute \(\mathbb{E}[D \,\mid\, G = g, A = a, I = i, M = m]\) to capture the average behavior under each condition.
\end{enumerate}

This serves as our \textit{baseline distribution} for later comparisons and manipulations.

\subsection{IV Steering Vectors} \label{sec:iv_steering_vectors}

\subsubsection{Basic Steering Vector} \label{sec:basic_steering_vector}

Section \ref{sec:vector_variable_variation} already provides a detailed explanation, here we briefly frame the key concepts within the Dictator Game context. We are interested in how certain IVs, such as gender (\(G\)), age (\(A\)), game instruction (\(I\)), or meeting condition (\(M\)), influence the final decision \(D\). To capture each IV's direction and magnitude of influence inside the LLM, we define a \textit{steering vector} as, consider an IV \(X \in \{G, A, I, M\}\) that can take two distinct values \(x_1\) and \(x_2\). Holding all other variables fixed or randomized\footnote{Whether other variables are held fixed or randomized, the resulting difference in the residual streams \(\mathbf{v}^{(\ell)}_{X}\) consistently captures the unique effect of the independent variable \(X\). When other variables are fixed, changing \(X\) directly isolates its impact on the residual stream. When randomized, any influence other variables might have is averaged out across multiple instances, neutralizing their impact.}, we measure the difference in the layer-\(\ell\) residual stream:

\[
    \mathbf{v}^{(\ell)}_{X}
    \,=\,
    \mathbf{R}^{(\ell)}(x_2)
    \;-\;
    \mathbf{R}^{(\ell)}(x_1).
\]

\(\mathbf{v}^{(\ell)}_{X}\) quantifies how much the internal representation at layer \(\ell\) shifts due to changing \(X\) from \(x_1\) to \(x_2\).

\subsubsection{Partial Steering Vector} \label{sec:partial_steering_vector}

In realistic setups, multiple IVs (e.g., \(G, A, I, M\)) interact in the prompt. For instance, the LLM might conflate \textit{female} with \textit{younger} if these often co-occur in training data. To focus on a single IV \(X_1\) independently, we remove the influence of other IVs \(X_2, X_3, \dots\).

For two variables \(X_1\) and \(X_2\), if \(\mathbf{v}^{(\ell)}_{X_1}\) and \(\mathbf{v}^{(\ell)}_{X_2}\) are their respective steering vectors, the \textit{partial} steering vector of \(X_1\) controlling for \(X_2\) is:

\[
    \mathbf{v}^{(\ell)}_{X_1 \mid X_2}
    \,=\,
    \mathbf{v}^{(\ell)}_{X_1}
    \;-\;
    \frac{\langle
        \mathbf{v}^{(\ell)}_{X_1},\,
        \mathbf{v}^{(\ell)}_{X_2}
        \rangle}
    {\|\mathbf{v}^{(\ell)}_{X_2}\|^2}
    \,\mathbf{v}^{(\ell)}_{X_2}.
\]

By subtracting the component of \(\mathbf{v}^{(\ell)}_{X_1}\) aligned with \(\mathbf{v}^{(\ell)}_{X_2}\), we isolate the ``pure'' effect of \(X_1\). This procedure can be repeated or extended to three or more variables using standard multivariate orthogonalization, generating a partial steering vector \(\mathbf{v}^{(\ell)}_{X \mid \{\text{others}\}}\).

For example, if \(X_1 = \text{gender (male vs. female)}\) and \(X_2 = \text{age (young vs. older)}\), \(\mathbf{v}^{(\ell)}_{X_1 \mid X_2}\) removes age-related correlations from the gender vector. Thus, we can specifically measure and manipulate ``male vs. female'' while \textit{not} shifting the representation in a direction correlated with ``young vs. older.''

\subsection{DV Steering Vector} \label{sec:dv_steering_vector}

Following the same rationale as for IVs, we can define a \textit{DV steering vector} that captures how the LLM's internal representation changes when the final decision \(D\) shifts from one anchor to another. In our Dictator Game, \(D\) is the amount of dollars allocated to the recipient, which can range from \(-20\) (taking all) to \(+20\) (giving all), depending on ``take'' vs. ``give'' frames. To identify the model's DV steering vector, we pick two typical reference decisions, \(D=10\) (a ``fair'' split of 20) and \(D=0\) (giving nothing). At layer \(\ell\):

\[
    \mathbf{v}^{(\ell)}_{D}
    \,=\,
    \mathbf{r}^{(\ell)}(x \,\mid\, D=10)
    \;-\;
    \mathbf{r}^{(\ell)}(x \,\mid\, D=0).
\]

\(\mathbf{v}^{(\ell)}_{D}\) is how the residual stream changes when the model's output changes from \(D=0\) to \(D=10\) at layer \(\ell\).

\subsection{Projecting IV Steering Vector onto DV} \label{sec:projection_partial_iv_steering}

The next question is how strongly each IV pushes the LLM's decision \(D\). To see whether a partial steering vector \(\mathbf{v}^{(\ell)}_{X \mid \{\text{others}\}}\) (for an IV \(X\)) actually impacts the decision \(D\), we project it onto \(\mathbf{v}^{(\ell)}_{D}\):

\[
    \text{Proj}_{D}
    \Bigl(\mathbf{v}^{(\ell)}_{X \mid \{\text{others}\}}\Bigr)
    \,=\,
    \frac{
    \bigl\langle
    \mathbf{v}^{(\ell)}_{X \mid \{\text{others}\}},\,
    \mathbf{v}^{(\ell)}_{D}
    \bigr\rangle
    }
    {\|\mathbf{v}^{(\ell)}_{D}\|^2}
    \,\mathbf{v}^{(\ell)}_{D}.
\]

The \textit{sign} of \(\langle \mathbf{v}^{(\ell)}_{X \mid \{\text{others}\}},\, \mathbf{v}^{(\ell)}_{D}\rangle\) indicates whether \(X\) \textit{increases} or \textit{decreases} \(D\). The \textit{magnitude} of this projection (\(\|\text{Proj}_{D}(\cdot)\|\)) is an ``effect size'': a larger value suggests that toggling \(X\) can produce a bigger shift between \(D=0\) and \(D=10\).

For example, if \(\mathbf{v}^{(\ell)}_{G=\text{female} \mid \{A, I, M\}}\) has a positive projection on \(\mathbf{v}^{(\ell)}_{D}\), it implies that, all else equal, adopting a ``female'' persona pushes the model to giving more dollars. Conversely, if \(\mathbf{v}^{(\ell)}_{A=60 \mid \{G, I, M\}}\) has a strong negative projection, it indicate that an older persona correlates with less generous allocations in the model's internal reasoning.

\subsection{Manipulating the Impact of IVs on DV} \label{sec:precise_control_ivs}

Having identified how strongly each IV influences \(D\), we now want to precisely manipulate the LLM's decision by variable. The key operation is to inject a vector into the residual streams such that it pushes the model in the direction of a higher (or lower) \(D\).

\subsubsection{Choice between Injecting IV Partial Steering Vector or Its Projection} \label{sec:choice_injecting_partial_iv}

The \textit{partial steering vector} for an IV \(X\), \(\mathbf{v}^{(\ell)}_{X \mid \{\text{others}\}}\), captures how changing \(X\) shifts the internal representation at layer \(\ell\), after orthogonalizing out the influence of other variables. The \textit{projection} of that partial vector \textit{onto the DV direction}, i.e. \(\text{Proj}_{D}\!\bigl(\mathbf{v}^{(\ell)}_{X \mid \{\text{others}\}}\bigr)\), is strictly the component of \(\mathbf{v}^{(\ell)}_{X \mid \{\text{others}\}}\) that points in the same (or opposite) direction as \(\mathbf{v}^{(\ell)}_{D}\).

If we inject the \textit{entire} partial steering vector, we might also be adding directions that do not project onto \(\mathbf{v}^{(\ell)}_{D}\). Those extra directions could have unintended effects on the model's reasoning or how it articulates its choice, possibly influencing style, justification, or other textual features that are not directly about increasing or decreasing \(D\). By contrast, if we only inject the \textit{projection}, we only push the model in the direction that moves \(D\), allowing for a more precise intervention.

Depending on the objective, either approach can be informative, and exploring the differences between the two can itself be an interesting, though potentially complex, research effort. In this pioneering work, our main priority is to demonstrate that manipulating the IVs can indeed produce expected changes in the model's decision. Consequently, to favor precision over comprehensiveness, we choose to inject \textit{only the projections} of the partial IV steering vectors.

\subsubsection{Injecting the Projection of Partial IV Steering Vector to Residual Stream} \label{sec:injecting_projection_partial_iv}

Let \(\alpha \in \mathbb{R}\) be an \textit{injection coefficient} that controls both the sign and magnitude of the intervention. For each layer \(\ell\), we replace the original residual stream \(\mathbf{r}^{(\ell)}(x)\) with

\[
    \widetilde{\mathbf{r}}^{(\ell)}(x)
    =
    \mathbf{r}^{(\ell)}(x)
    \;+\;
    \alpha \cdot
    \;\text{Proj}_{D}\!\bigl(\mathbf{v}^{(\ell)}_{X \mid \{\text{others}\}}\bigr).
\]

As a result, the pure representation of \(X\) in residual streams can be amplified when \(\alpha>0\) and reduced if \(\alpha<0\). Because we used the projection of the partial steering vector of \(X\), this manipulation remains specific to \(X\). That is, we are not inadvertently pushing the hidden state in directions aligned with other variables (like age or meeting condition).

\subsubsection{Changing LLM's Decision} \label{sec:changing_llm_decision}

Consider the function \(f\) that maps the final-layer residual stream \(\mathbf{r}^{(L)}\) to the decision variable \(D\), representing the original decision before any manipulation:

\[
    f(\mathbf{r}^{(L)}) = D.
\]

Let \(\mathbf{p}^{(L)}_{D} = \text{Proj}_{D}\!\bigl(\mathbf{v}^{(L)}_{X \mid \{\text{others}\}}\bigr)\). After introducing a small perturbation to the residual stream, we approximate the new decision as:

\[
    \begin{aligned}
        f(\widetilde{\mathbf{r}}^{(L)})
         & =
        f\Bigl(\mathbf{r}^{(L)} + \alpha \cdot \mathbf{p}^{(L)}_{D}\Bigr)                                                                       \\
         & \;\approx\;
        f(\mathbf{r}^{(L)}) + \alpha \cdot \Bigl\langle \nabla_{\mathbf{r}} f\bigl(\mathbf{r}^{(L)}\bigr), \; \mathbf{p}^{(L)}_{D} \Bigr\rangle \\
         & =
        D + \alpha \cdot \Bigl\langle \nabla_{\mathbf{r}} f\bigl(\mathbf{r}^{(L)}\bigr), \; \mathbf{p}^{(L)}_{D} \Bigr\rangle.
    \end{aligned}
\]

In this equation:

\begin{itemize}[topsep=0pt,itemsep=0ex,partopsep=0ex,parsep=0ex]
    \item \(\nabla_{\mathbf{r}} f(\mathbf{r}^{(L)})\): The \textit{gradient} of the output function with respect to the residual stream. This gradient indicates how sensitive the decision \(D\) is to changes in each component of \(\mathbf{r}^{(L)}\).
    \item \(\Bigl\langle \nabla_{\mathbf{r}} f\bigl(\mathbf{r}^{(L)}\bigr), \; \mathbf{p}^{(L)}_{D} \Bigr\rangle\): The dot product quantifies the \textit{direction and strength} of how \(\mathbf{p}^{(L)}_{D}\) affects \(D\).
    \item \(\alpha\): The injection coefficient that controls both the \emph{magnitude} and \emph{direction} of the perturbation.
\end{itemize}

Layer-by-layer adjustments compound in a similar manner due to the residual nature of the transformer architecture, meaning that interventions at different layers can have cumulative or localized effects on the final decision.

\subsection{Fine-Tuning the Manipulation} \label{sec:fine_tuning_manipulation}

Eventually, the final intervened decision depends on the product \(\alpha \cdot \Bigl\langle \nabla_{\mathbf{r}} f\bigl(\mathbf{r}^{(L)}\bigr), \; \mathbf{p}^{(L)}_{D} \Bigr\rangle\), where most of the terms are intrinsic to the model and the prompt. However, two key parameters can be fine-tuned to alter model decisions: the layer number \(L\) and the injection coefficient \(\alpha\).

\paragraph{Layer Number \(L\).} Determines the stage at which the residual stream is modified within the model. Intervening at \textit{earlier layers} can have a more \textit{amplified effect} on the final decision due to the cumulative nature of residual connections, while intervening at \textit{later layers} allows for more \textit{localized control} over specific aspects of the decision-making process.

\paragraph{Injection Coefficient \(\alpha\).} Controls the strength of the manipulation. A larger \(|\alpha|\) results in a more significant shift in the decision \(D\), either amplifying or reversing the influence of the independent variable \(X\). Existing studies suggest that \(|\alpha|\) should be smaller than 15 to generate meaningful and interpretable results; otherwise, the model may collapse and generate nonsensical outputs \autocite{TurnerSteeringLanguageModels2024}. We expand the range of \(\alpha\) to 30 to explore a larger spectrum of possible effects.

By adjusting these two parameters, we can further fine-tune the extent to which the model's decision is altered. Through the two parameters, our interventions can be both \textit{targeted} (through the choice of layer) and \textit{controllable} (through the scaling factor \(\alpha\)), enabling nuanced manipulation of the LLM's behavior.

\section{Results}

\subsection{Baseline: Model Performance and Behavior}

To ensure the LLM-agent behaves logically, we also include a verification question at the end of each trial to assess whether the agent accurately calculates the final allocations for both the dictator and the recipient after the transfer. Performance on this question can serve as a measure of the agent's mathematical reasoning and understanding of the game rules.

Out of 1,000 baseline trials, the agent produces logically correct responses in 571 instances (57.1\%). The giving rates exhibit a bimodal distribution, with agents either giving nothing (200 trials, 35.03\%) or giving half (371 trials, 64.97\%). Compared to the findings of \textcite[19-20]{MaCanMachinesThink2024}, where only 40.43\% of trials are logically correct and more than half of these involve giving nothing, our experiment demonstrates both improved logical accuracy and greater generosity. However, due to the simplified game setup in this study, the two results may not be directly comparable. Table \ref{tab:baseline_distribution} and Figure \ref{fig:age_distribution} in the appendix show the distribution of all variables.

Table \ref{tab:logit_coefficients} presents the logistic regression results predicting the likelihood of transferring a non-zero amount in the Dictator Game. The DV is binary, with 0 indicating no transfer and 1 indicating a transfer of a non-zero amount. The coefficient for \textit{Give Framing} is 1.059 (\(p < .001\)), indicating that agents exposed to a ``give'' framing are approximately 2.88 times (\(\approx e^{1.059}\)) more likely to transfer a non-zero amount compared to those exposed to a ``take'' framing. This finding, while seemingly straightforward, provides a crucial validation of our experimental setup. The fact that the model is highly sensitive to this fundamental framing of the task confirms that it has learned the basic structure of the Dictator Game and is responding in a way that is consistent with the experimental manipulations. This serves as a critical baseline, demonstrating that the model is ``paying attention'' to the experimental conditions, which gives us confidence in the more nuanced findings that follow. Similarly, the coefficient for \textit{Stranger Meet} is 0.815 (\(p < .001\)), suggesting that dictators are about 2.26 times (\(\approx e^{0.815}\)) more likely to make a non-zero transfer to recipients if they meet after the game, compared to having no interaction at all.

\begin{table}[htbp]
    \centering
    \caption{\textsc{Baseline: Logistic Regression Predicting Amount Transferred}}
    \label{tab:logit_coefficients}
    \begin{threeparttable}
        \begin{tabular}{@{}r@{\hspace{20pt}}r@{\hspace{20pt}}r@{\hspace{20pt}}r@{}}
            \hline\hline
            Variable      & Coef. & 95\% CI        & $p$     \\
            \hline
            Give Framing  & 1.059 & [.696, 1.423]  & $<.001$ \\
            Stranger Meet & .815  & [.439, 1.192]  & $<.001$ \\
            Female        & .211  & [-.153, .575]  & .256    \\
            Age           & .001  & [-.015, .017]  & .901    \\
            Intercept     & .640  & [-.046, 1.326] & .068    \\
            \hline\hline
        \end{tabular}
        \begin{tablenotes}[para,flushleft]
            \footnotesize
            \textit{Notes}: $N=571$; Pseudo $R^{2} = 0.077$. Dependent variable: Amount Transferred Binary (0 for nothing transferred, 1 for non-zero amount transferred). CI = Confidence Interval.
        \end{tablenotes}
    \end{threeparttable}
\end{table}

The variable \textit{Female} has a positive but non-significant coefficient (0.211, \(p = .256\)), suggesting that LLM-agents with a ``female'' persona may be 1.24 times (\(\approx e^{0.211}\)) more likely to transfer a non-zero amount than ``male'' agents; however, this effect is not statistically significant. Similarly, \textit{Age} shows a negligible effect on transfer likelihood (coefficient: 0.001, \(p = .901\)).

The intercept coefficient (0.640, \(p = .068\)) reflects the baseline log odds of transferring a non-zero amount, with an odds ratio of 1.90 (\(\approx e^{0.640}\)). This indicates that LLM-agents are prosocial by default, being nearly twice as likely to transfer a non-zero amount than nothing. Overall, the model accounts for a modest proportion of variance in the dependent variable (Pseudo \(R^2 = 0.077\)).

\subsection{Baseline: Computational Basis of Variables}

\begin{sidewaysfigure}[htbp]
    \centering
    \caption{\textsc{Relations between IV and DV steering vectors}} \label{fig:relations_IV_DV_steering_vectors}
    \includegraphics[width=1\textwidth]{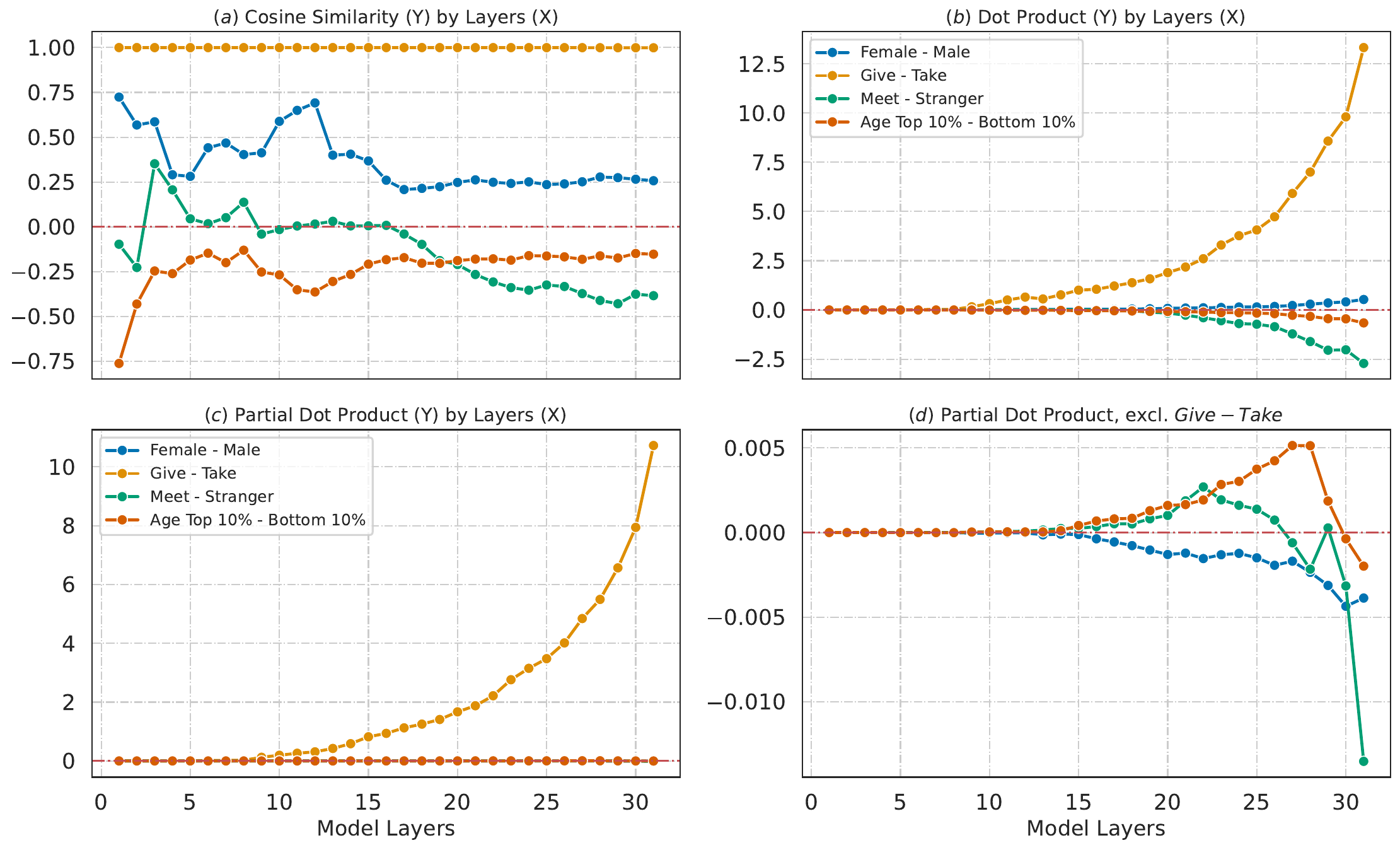}
    \begin{minipage}{1\textwidth}
        \footnotesize
        \textit{Notes}: (\textit{d}) is the same as (\textit{c}) but with $Give-Take$ excluded to better visualize the other variables. These profiles characterize internal relations in LLMs' residual streams and should not be confused with human cognition or behavior.
    \end{minipage}
\end{sidewaysfigure}

The descriptive and regression results provide a behavioral evaluation of the LLM's decision-making process. To explore its internal mechanisms, we examine the representations of IVs, DVs, and their relationships through LLM residual streams. Figure \ref{fig:relations_IV_DV_steering_vectors} illustrates the relationship between IV steering vectors and the final decision vector across all layers of the LLM model. Be reminded that these relationships index within-model geometry (alignment and magnitude) rather than human-level construct validity.

\begin{enumerate}[topsep=0pt,itemsep=0ex,partopsep=0ex,parsep=0ex,label=(\alph*)]
    \item \textit{Cosine similarity} (Figure \ref{fig:relations_IV_DV_steering_vectors}(\textit{a})) measures the alignment between IV and DV steering vectors, indicating the degree to which they are directionally consistent within the LLM's internal semantic space. High cosine similarity values (close to 1) imply a strong directional alignment, signifying that the IV has a significant positive influence on the decision-making process. Conversely, values close to -1 indicate that the vectors point in exactly opposite directions, meaning the IV has an inverse influence on the decision. Values near 0 suggest minimal or no directional influence of the IV on the decision.
    \item \textit{Dot product} (Figure \ref{fig:relations_IV_DV_steering_vectors}(\textit{b})) quantifies the overall strength of each IV's influence on the decision vector. Higher dot product values indicate a greater magnitude of influence, reinforcing the observations from cosine similarity regarding which IVs are more impactful.
    \item \textit{Partial dot product} (Figures \ref{fig:relations_IV_DV_steering_vectors}(\textit{c}) and \ref{fig:relations_IV_DV_steering_vectors}(\textit{d})), by isolating the unique contribution of each IV, reveals how much each variable independently affects outputs, controlling for the influence of other IVs.
\end{enumerate}

\subsubsection{Cosine Similarly: Alignment between IV and DV Steering Vectors}

Figure \ref{fig:relations_IV_DV_steering_vectors}(\textit{a}) presents the cosine similarity values between IV and DV steering vectors across all layers of the LLM. The results reveal several key patterns:

\begin{enumerate}[topsep=0pt,itemsep=0ex,partopsep=0ex,parsep=0ex,label=(\arabic*)]
    \item Consistent Directional Alignment: The directional alignment between IV and DV steering vectors remains relatively consistent across different layers of the language model. This suggests that, at each stage of processing, the model's internal semantic space represents these variables and their relationships in a coherent and stable manner.
    \item Layer-Specific Variations: The alignments exhibit more variations in the earlier layers (Layers 1--20) but stabilize in the later layers (Layers 20--31). This indicates that initial processing stages are more sensitive to the influence of IVs, while deeper layers refine these representations, resulting in more stable and consistent alignments with the decision-making process.
    \item Dominant Influence of Framing: The $Give-Take$ framing exhibits the highest cosine similarity (close to 1) with the DV steering vector, indicating that this variable has the most consistent and strongest directional influence on the model's decision-making process.
    \item Moderate Influence of Gender: The $Female$ variable shows a moderate cosine similarity (ranging between 0.75 and 0.25) with the DV steering vector. This suggests that gender information in the LLM's internal representation has a discernible but less dominant impact on the final decision compared to framing.
    \item Evolving Influence of Meeting Condition: The cosine similarity of the $Meet-Stranger$ variable with the DV steering vector is low in the early layers but gradually increases in the later layers. This indicates that the model's representation of meeting conditions becomes more aligned with the decision-making process as it processes information through deeper layers.
    \item Diminishing Influence of Age: The $Age$ variable exhibits high directional alignment with the DV steering vector in the first two layers but sharply decreases to relatively low and gradually decreasing levels across the later layers. This suggests that age information plays a significant role in the initial stages of decision-making but becomes less influential as processing progresses.
\end{enumerate}

In general, we observe that certain IVs, particularly those related to the framing of the game and demographic attributes, exert varying levels of influence on the decision-making process across different layers of the LLM. The stronger alignment of framing variables like $Give-Take$ underscores their pivotal role in guiding the model's decisions, while demographic factors like gender and age exhibit more nuanced influences that evolve through the model's layers.

\subsubsection{Dot Product and Partial Dot Product: Magnitude of IV Influence}

The results of the dot product and partial dot product, presented in Figures \ref{fig:relations_IV_DV_steering_vectors}(\textit{b}), \ref{fig:relations_IV_DV_steering_vectors}(\textit{c}), and \ref{fig:relations_IV_DV_steering_vectors}(\textit{d}), provide additional insights into the magnitude of each IV's influence on the decision vector and the unique contribution of each IV to the decision-making process. Specifically:

\begin{enumerate}[topsep=0pt,itemsep=0ex,partopsep=0ex,label=(\arabic*)]
    \item Cumulative Influence Across Layers: The magnitude of both the dot product and partial dot product increases progressively across the layers. This indicates that the influence of IVs accumulates as information flows through the model's layers, leveraging the residual connections to amplify their impact on the final decision.
    \item Dominant Influence of Framing: The influence of the $Give-Take$ framing is significantly higher than other variables when measured by the dot product. Even after controlling for the influence of other variables, as shown by the partial dot product, the $Give-Take$ framing remains the highest. In contrast, the influence of other variables becomes minimal, highlighting the dominant role of framing in LLM's decision-making process.
\end{enumerate}

\subsubsection{Dissociation Between Alignment and Magnitude}

Our analysis of the computational basis of variables reveals a noticeable dissociation between the alignment (cosine similarity) and the magnitude (dot product and partial dot product) of IV influence. Specifically, the directional alignment of IVs with the DV does not always correspond to their impact on final outcomes. For example, the gender persona moderately aligns (i.e., cosine similarity) with the decision vector (Figure \ref{fig:relations_IV_DV_steering_vectors}(\textit{a}), blue line), but its final impact (i.e., dot product) on decision is miniscule (Figure \ref{fig:relations_IV_DV_steering_vectors}(\textit{d}), blue line). This indicates that both the direction and magnitude of IV steering vectors play roles in shaping the decision-making process.

\subsection{Manipulation Analysis}

\subsubsection{Model Statistics of Manipulations} \label{sec:model_statistics_of_manipulations}

We conducted a total of 1,891 manipulations (61 injection coefficients $\times$ 31 layers,\footnote{We inject at 31 internal layers (Layers 1--31), excluding the final decoder layer ($L=32$) to avoid direct perturbations at the logits head.} see Section \ref{sec:fine_tuning_manipulation}). For each manipulation, we performed 1,000 trials and subsequently conducted regression analyses using only the logically correct trials to obtain the regression coefficients for all IVs and overall model statistics (see an example manipulation with $\alpha = 30$ and $\ell=1$ annotated in Figure \ref{fig:rc_heatmap_female}).

\begin{figure}[htbp]
    \centering
    \caption{\textsc{Histograms of manipulation-level metrics}}
    \label{fig:model_stats_hist}
    \includegraphics[width=1\textwidth]{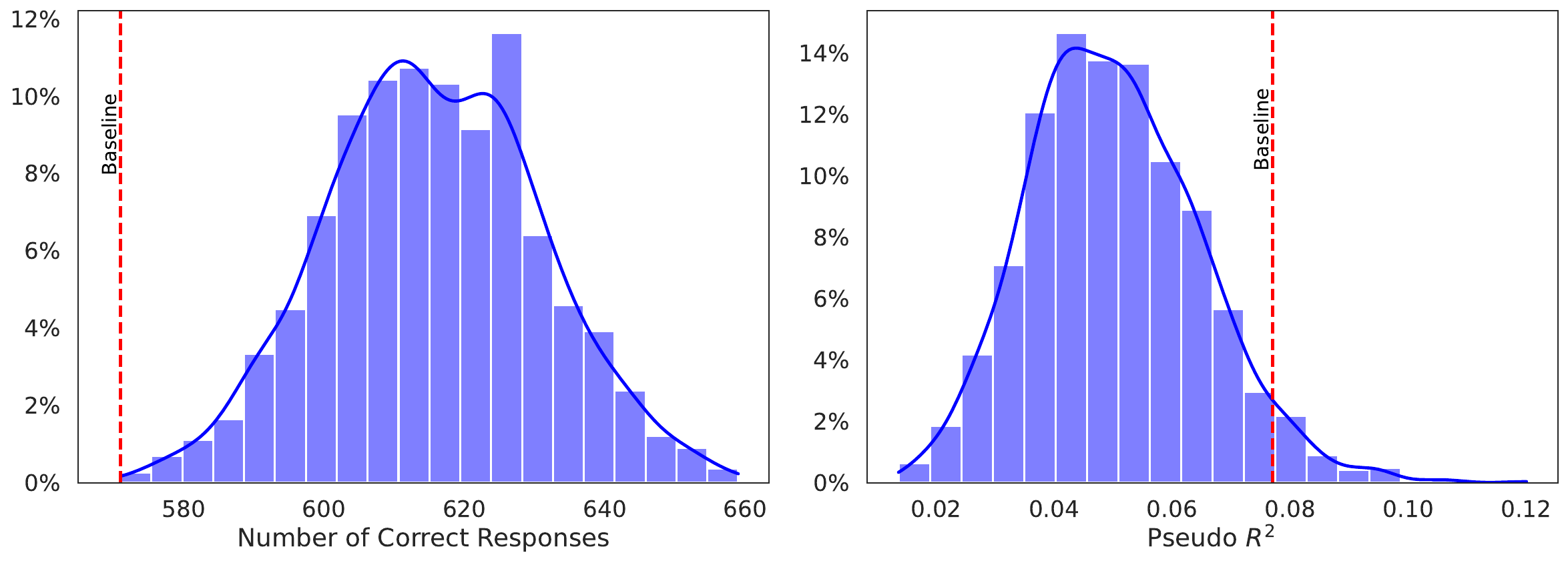}
    \begin{minipage}{1\textwidth}
        \footnotesize
        \textit{Notes}: Left: count of logic-check-passing trials (data-quality metric). Right: Pseudo-$R^{2}$ from the logistic model predicting non-zero transfer using the IVs (i.e., Give/Take, Meet/Stranger, Female/Male, Age). Higher left values = more usable trials; higher right values = stronger IV$\rightarrow$DV association conditional on the IVs.
    \end{minipage}
\end{figure}

Figure \ref{fig:model_stats_hist} shows the distributions of manipulation-level metrics. The number of logic-check-passing responses (left) is approximately normal, ranging from 571 to 659 ($Mean = 615.47$, $SD = 15.33$). Pseudo-$R^2$ (right) is also approximately normal, ranging from 0.014 to 0.120 ($Mean = 0.050$, $SD = 0.015$). Relative to baseline (Pseudo-$R^2=0.077$ with 571 usable trials), manipulations, on average, increase the share of logic-check-passing trials ($z=2.9$) while lowering Pseudo-$R^2$ ($z=1.8$).

\emph{Why do more trials pass the logic check after injection?} Our intervention gently ``nudges'' the model in the direction of making a clearer decision. This extra nudge helps it stick to the game's standard response format and do the simple arithmetic correctly. With fewer off‑template explanations or calculation slips, more trials pass the logic check.

\emph{Why does Pseudo-$R^2$ go down?} The regression only uses the listed variables (Give/Take, Meet/Stranger, Female/Male, Age) to explain whether the model makes any transfer. Our intervention adds an additional push toward deciding that is not recorded in those variables. Because part of the model’s behavior now comes from this extra push (not from the IVs), the IVs by themselves explain a smaller share of the variation in outcomes. That makes Pseudo-$R^2$ lower.

\emph{Why can both happen at the same time?} These two metrics measure different things. The logic check is a data‑quality test (did the model follow the rules and arithmetic?), while Pseudo-$R^2$ asks how much of the outcome is explained by the IVs alone. A small decision‑stabilizing nudge can make outputs cleaner and more consistent (raising pass rates) while also shifting some of the decision to something outside the IVs (lowering the amount the IVs can explain). Thus, better data quality and lower Pseudo-$R^2$ can coexist without contradiction.

\subsubsection{Regression Coefficients of Steered ``$Female$''}

\begin{sidewaysfigure}[htbp]
    \centering
    \caption{\textsc{Regression Coefficients of Steered ${Female}$}} \label{fig:rc_heatmap_female}
    \includegraphics[width=1\textwidth]{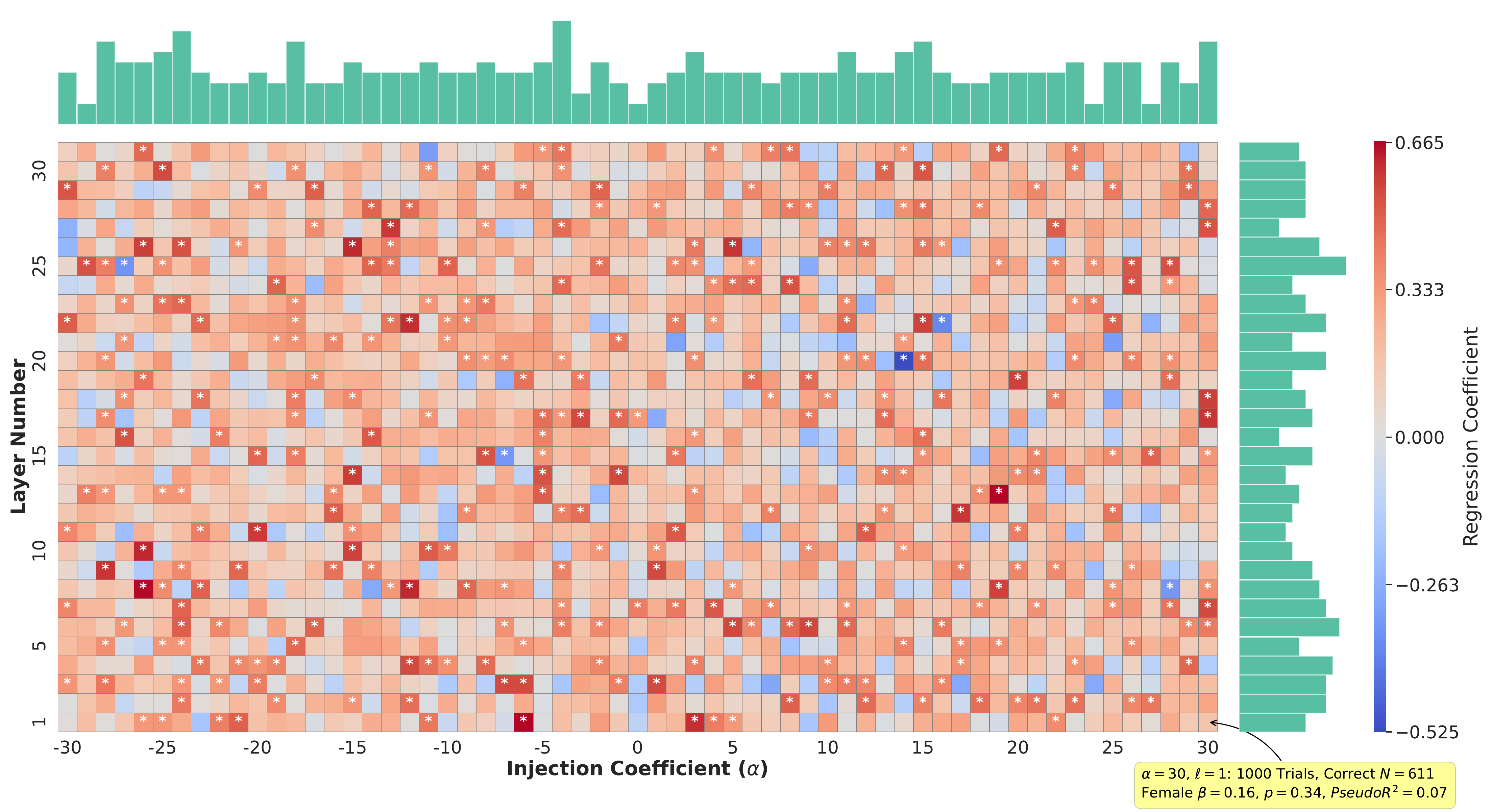}
    \begin{minipage}{1\textwidth}
        \footnotesize
        \textit{Notes}: Each cell represents the regression coefficient of the $Female$ variable for a specific combination of layer number and injection coefficient. Cells marked with asterisks indicate coefficients that are statistically significant at the $p < 0.05$ level. The marginal histograms show the number of significant coefficients (i.e., the number of asterisks) by layer or injection coefficient.
    \end{minipage}
\end{sidewaysfigure}

Figure \ref{fig:rc_heatmap_female} displays the regression coefficients of the steered $Female$ variable across all layers and injection coefficients. In this figure, each cell represents the regression coefficient of the $Female$ variable for a specific combination of layer number and injection coefficient (with the cell annotated for $\alpha = 30$ and $\ell=1$ as an example). Regression coefficients that are statistically significant at the $p < 0.05$ level are marked with asterisks.

Among all 1,891 manipulations, 320 (16.92\%) resulted in statistically significant regression coefficients for the $Female$ variable. Of these significant coefficients, 315 (98.44\%) are positive, ranging continuously from 0.32 to 0.66. This indicates that a majority of the manipulations lead to an increase in the likelihood of transferring a non-zero amount even when the injection coefficients are negative (meaning steered \textit{away} from ``female''). Since these coefficients span a continuous range, as Figure \ref{fig:rc_heatmap_female_ordered} illustrates, it is possible to select a specific manipulation to achieve a desired effect size.

The analysis of the steered $Female$ variable reveals that a substantial proportion of manipulations significantly influence the model's decision-making process, aligning with our theoretical expectations. However, these manipulations do not exhibit clear patterns, and the majority of significant coefficients remain positive, even when the injection coefficient is negative. Theoretically, we would expect a stronger positive correlation between the injection coefficient and the resulting regression coefficient, but the observed correlation is minuscule ($\lvert r\rvert < 0.05$). These findings suggest that the model's internal representations involve complex interactions that are not fully captured by the current manipulation framework.

\subsection{Orthogonality Analysis}

According to our theoretical framework, the injected IV steering vectors are orthogonal to one another. This orthogonality implies that the effects of manipulating different IVs should be independent. Specifically, the manipulation of the $Female$ variable should not influence the regression coefficients of other IVs.

Figures \ref{fig:coef_diff_heatmap_meet} to \ref{fig:coef_diff_heatmap_give} in the appendix present the orthogonality analysis. The regression coefficients of the $Age$ and $Give$-$Take$ variables are rarely significantly affected by the manipulation of the $Female$ variable. This suggests that the steering vectors of different IVs are largely orthogonal, supporting the independence of manipulations.

However, the coefficient of the $Meet$-$Stranger$ variable is more frequently influenced by the manipulation of the $Female$ variable, indicating a potential interaction between these two variables. In human studies, multiple lines of research show that women tend to engage in more communal, relational forms of prosocial behavior, whereas men often exhibit more agentic, strength-intensive, or collectively oriented actions \autocite{EaglyHisHersProsocial2009}. Moreover, framing manipulations in dictator games suggest that women often showing higher generosity toward strangers than men do in certain contexts \autocites{ChowdhuryGenderDifferencesGiving2017}{ChowdhuryGenderDifferencesGiving2020}. Similarly, as social distance increases, women's generosity changes significantly more than men's \autocite{Donate-BuendiaGenderOtherModerators2022}. Altogether, these findings align with our observation that $Meet$-$Stranger$ is more sensitive to $Female$ manipulations, suggesting that the model may have captured the documented gender-context interplay.

Eventually, we still observe 284 robust manipulations (i.e., manipulations in which only the $Female$ variable is significantly affected but not the other IVs), a substantial proportion of the successful manipulations of $Female$ (284 out of 320, or 88.75\%). This high rate of orthogonal manipulation implies that, despite some interaction between $Female$ and $Meet$-$Stranger$, the steering vectors remain largely independent in most cases. Theoretically, this underscores that gender-related shifts often intersect with social-context cues but can still be distinctly controlled in activation steering. Practically, it highlights the viability of using separate steering vectors to manipulate specific variables without extensively ``spilling over'' into others, an important consideration for applications requiring precise control over multiple attributes.

\clearpage
\section{Discussion}

LLMs occupy a dual theoretical position in sociological research: they can be (i) repositories of culturally structured meanings acquired through large‐scale training, and (ii) adaptive agents whose outputs can be experimentally probed as if they were synthetic participants \autocites{ZiemsCanLargeLanguage2024}{BailCanGenerativeAI2024}{AnthisLLMSocialSimulations2025}. These roles raise a core methodological challenge reviewed earlier: prevailing validation strategies (prompt engineering, output benchmarking, post‑hoc statistical correction) treat models as opaque input-output devices, leaving unanswered whether their behavior reflects robust internal models of social meaning or merely brittle pattern matching \autocites{AherUsingLargeLanguage2023}{MaCanMachinesThink2024}{KozlowskiSimulatingSubjectsPromise2025}. Theoretical leverage for sociology---linking cultural schemas, role expectations, and framing effects to behavior---requires opening that black box to identify where and how socially meaningful distinctions (e.g., gender, age, and moral frames) reside in internal activation space.

Our results operationalize this theoretical program: if social meanings are instantiated as approximately linear directions in the residual stream \autocites{ParkLinearRepresentationHypothesis2024}{TiggesLinearRepresentationsSentiment2023}, then they can be (1) estimated via experimental contrasts, (2) disentangled via orthogonalization, and (3) causally manipulated through controlled injection. This reframes LLMs from inscrutable statistical artifacts into quantifiable internal representations whose geometry (angles, projections, magnitudes) encodes sociological variables, analogous to how network analysis formalized relational structure decades earlier \autocites{Arseniev-KoehlerTheoreticalFoundationsLimits2024}{BorgattiNetworkAnalysisSocial2009}. Activation engineering thus supplies a bridge between theories of meaning and computational implementation: it converts abstract constructs into manipulable vectors, enabling virtual experiments on internal representations rather than only surface responses.

This study introduces and validates a method for probing and steering internal representations of social concepts in LLMs. Combining classic experimental design with activation engineering, we isolate, measure, and manipulate the influence of specific social variables on model behavior, showing that variables are encoded with varying strength and depth and that steering can orthogonally control their impact on decisions. Methodologically, this contributes (1) an analytical strategy that moves beyond treating LLMs as ``black boxes,'' replacing sole reliance on prompt-based validation with direct inspection of internal processing for more rigorous, theory-grounded assessment of simulations, and (2) a practical tool for computational social scientists and industry applications: by operationalizing social concepts as vectors, researchers can run virtual experiments to test causal hypotheses in controlled settings, expanding theory building where conventional experiments are difficult or impossible.

\label{par:add_clarity}
For additional conceptual clarity and accuracy, our contribution concerns \emph{internal representational validity} and controllability of LLMs, where social variables reside in activation space and how they can be causally steered. We do not claim \emph{external sociological validity}; linking these internal directions to human cognition and behaviors is a separate empirical question requiring comparative studies.

\subsection{Research Design Guidelines for Social Scientists} \label{sec:research_design_guidelines}

Activation-based steering can be treated as a research instrument rather than a black-box trick. It enables three concrete uses. First, \emph{validity audits}: assess whether core constructs (e.g., social distance, gendered prosociality, fairness) are robustly encoded as internal directions; if a construct is weakly represented, simulations relying on it are unlikely to be theoretically and empirically faithful. Second, \emph{theory-driven experiments}: inject small, controlled perturbations along a construct's direction to test hypotheses (e.g., whether gender cues increase prosocial transfers), identify boundary conditions, and generate new hypotheses. Third, \emph{measurement}: compare construct directions across models and checkpoints to track ``cultural drift'' in what is encoded. These practices interrogate internal mechanisms and should complement, never replace, studies with human participants.

\subsubsection{Key Steps: Variable-Specific Steering in Social Simulations with LLMs} \label{sec:key_steps}

The goal is precise, variable-specific control over LLM behavior so that simulated responses can align with observed social patterns of humans, achieving a higher degree of external validity. The research-design framing below aligns LLM simulation with standard social science practices: clear constructs, pre-specified contrasts, conservative interventions, human benchmarking, transparency, and reproducibility.

\begin{enumerate}[topsep=0pt,itemsep=0ex,partopsep=0ex,parsep=0ex,label=(\arabic*)]
    \item \emph{Clarify constructs and operationalizations.} Define the conceptual variables of interest (e.g., gender persona, social distance, framing) and link each to concrete prompt fields and outcome definitions. Pre-specify value ranges and coding rules for both inputs and the dependent variable.
    \item \emph{Establish a baseline.} Run a representative or random set of prompts spanning the operational variables. Record both behavioral outcomes and the internal signals needed for later analysis. This baseline functions as your ``unmanipulated'' distribution against which all interventions are judged.
    \item \emph{Isolate variables.} Disentangle overlapping constructs so that each variable’s effect can be examined on its own. Conceptually, you are removing shared variation to focus on the unique component of each variable. This supports interpretable, variable-specific tests.
    \item \emph{Pre-specify interventions and success criteria.} Select layers to test and a conservative grid of intervention strengths. Start small and define \textit{ex-ante} success metrics such as expected direction of change on the outcome, acceptable collateral movement on non-target variables, and minimal disruption to response quality.
    \item \emph{Run within-model experiments.} Inject the variable-specific signal during inference to test causal hypotheses inside the model. Keep prompts and decoding settings constant, and change one factor at a time to preserve experimental control.
    \item \emph{Benchmark against human data.} Triangulate manipulated outcomes with human benchmarks (e.g., pre-registered replications or archival datasets). Choose manipulations that yield the closest alignment, transparently report mismatches, and treat any alignment as an approximate calibration, not proof of equivalence, due to external validity concerns (Section \ref{sec:llm_opportunities_challenges}).
    \item \emph{Documentation and governance.} Preregister contrasts and evaluation criteria where possible; release prompts, random seeds, LLM parameters, and intervention logs to support reproducibility. Note ethical safeguards (e.g., bounding interventions along sensitive demographic axes) and intended use constraints.
\end{enumerate}

\subsubsection{Example Modular Code Scripts} \label{sec:example_script}

In addition to these guidelines, we released three core code scripts\footnote{\url{https://doi.org/10.17605/OSF.IO/J9RQK}} used in this study to facilitate verification, further research, and application of variable-specific steering methods. These scripts are modular and can be adapted for different models and experiments.

\begin{itemize}[topsep=0pt,itemsep=0ex,partopsep=0ex,parsep=0ex]
    \item \texttt{core\_functions.py}: Contains all essential functions for model manipulation and analysis.
    \item \texttt{play\_dictator\_baseline\_residuals.py}: Implements the baseline experimental setup for measuring residuals.
    \item \texttt{play\_dictator\_steer\_amount\_gender\_partial.py}: Conducts experiments on gender steering effects.
\end{itemize}

\subsection{Industry Applications} \label{sec:industry_applications}

Overall, any entity that wants flexible, precise governance over LLM outputs---while reducing the overhead of constant retraining---may find the method introduced in this study valuable. For example, it can be applied to encourage prosocial behavior in AI-driven systems by selectively amplifying or diminishing certain variables, thereby promoting fairness or cooperation.

In marketing research, this method enables the creation of more valid social simulations by populating them with LLM agents whose behaviors can be manipulated. Researchers can construct virtual audience panels where latent traits like ``genre affinity,'' ``trailer responsiveness,'' or ``ticket-price sensitivity'' are operationalized as distinct, manipulable steering vectors. This allows for clean, virtual A/B tests that isolate the causal impact of specific marketing interventions. For example, one could predict a film’s opening-week theater viewing volume by varying the trailer cut or media mix while holding the ``ticket-price sensitivity'' vector constant, thereby eliminating confounding effects that often undermine prompt-only approaches. This technique facilitates rapid, low-cost testing of strategies (e.g., release timing, trailer variants, cast-focused creatives, geo-targeted spend) on large, customizable synthetic cohorts. The most promising scenarios can then be validated with smaller, targeted field experiments, such as limited-market screenings or geo-split ad campaigns using admissions as the outcome, reducing costs and accelerating the greenlight and marketing optimization cycle.

For research service platforms, providers of online experimentation and survey services (e.g., Qualtrics and Prolific) could adopt this method to run ``synthetic subject'' studies, in which factors like demographic attributes or question framing are precisely manipulated. This level of control enables novel experimental designs that would be challenging or impractical to conduct with human participants, offering researchers a more scalable and ethically flexible way to investigate complex behaviors.

In more entertainment-focused settings, role-playing and story-driven games can incorporate the method to generate richer, more responsive narratives. By injecting or subtracting specific steering vectors (e.g., ``cautious'' vs.\ ``bold'' traits), developers and players can fine-tune how AI-driven characters behave, speak, or make decisions, allowing for deeply personalized storylines.

These examples highlight just a few of the many possibilities for applying the method in real-world contexts. By offering transparent, fine-grained control over the underlying variables that guide an LLM's behavior, this study's approach stands to benefit not only social scientists but also commercial applications seeking to manage complex AI-driven interactions responsibly and efficiently.

\subsection{Future Directions} \label{sec:future_directions}

This study provides a foundational method for probing and steering social concepts within LLMs. To build on this foundation and move from a proof-of-concept to a robust tool for social science, future work could prioritize making the approach more powerful, accessible, and sociologically valid. We outline three key directions:

\emph{From Scripts to User-Friendly Tools.} To make activation-based steering more accessible, a crucial next step is to develop a user-friendly software package. Such a package would abstract away low-level implementation details by providing a high-level API, support for multiple models, comprehensive documentation, and built-in visualization tools, thereby increasing accessibility for a broader range of researchers.

\emph{Expanding to other Social Constructs and Experiments.} The framework developed in this study is highly generalizable. With a user-friendly software package, future research could explore its application to a wider range of social constructs and experimental paradigms, such as moral psychology, political science, and organizational behavior, to build a more comprehensive understanding of how LLMs represent the social world.

\emph{Bridging the Gap Between Internal and External Validity.} While this study focuses on the \textit{internal representational validity} of LLMs, future work must establish \textit{external sociological validity}. This requires systematic human-LLM comparisons, the use of activation-based steering for cognitive modeling, and large-scale validation studies to test if and how LLM behavior corresponds to that of human subjects.

\bigskip

\noindent Taken together, our results sketch a practical bridge between sociological theory and the internal geometry of modern language models. By working inside the model---measuring, comparing, and steering social meanings as directions in representation space---we turn LLMs from opaque black boxes into calibrated instruments for theory testing, bounded synthetic data, and auditable interventions. The program we outlined is intentionally modular: each step invites open benchmarks, uncertainty estimates, and guardrails that the community can scrutinize and improve. As these practices spread, debates about whether LLM agents ``simulate people'' will give way to cumulative evidence about which regularities they capture, where they fail, and how to align them with human values. The opportunity is not to replace judgment but to enhance it, making our models clearer, our inferences humbler, and our science faster.

\clearpage
\begingroup
\onehalfspacing
\sloppy
\printbibliography[title={References}]

@inproceedings{AherUsingLargeLanguage2023,
  title = {Using {{Large Language Models}} to {{Simulate Multiple Humans}} and {{Replicate Human Subject Studies}}},
  booktitle = {Proceedings of the 40th {{International Conference}} on {{Machine Learning}}},
  author = {Aher, Gati V. and Arriaga, Rosa I. and Kalai, Adam Tauman},
  date = {2023-07-03},
  pages = {337--371},
  publisher = {PMLR},
  issn = {2640-3498},
  url = {https://proceedings.mlr.press/v202/aher23a.html},
  urldate = {2024-10-05},
  eventtitle = {International {{Conference}} on {{Machine Learning}}},
  language = {en}
}

@online{AnthisLLMSocialSimulations2025,
  title = {{{LLM Social Simulations Are}} a {{Promising Research Method}}},
  author = {Anthis, Jacy Reese and Liu, Ryan and Richardson, Sean M. and Kozlowski, Austin C. and Koch, Bernard and Evans, James and Brynjolfsson, Erik and Bernstein, Michael},
  date = {2025-04-03},
  eprint = {2504.02234},
  eprinttype = {arXiv},
  eprintclass = {cs},
  doi = {10.48550/arXiv.2504.02234},
  pubstate = {prepublished}
}

@article{ArgyleOutOneMany2023,
  title = {Out of {{One}}, {{Many}}: {{Using Language Models}} to {{Simulate Human Samples}}},
  shorttitle = {Out of {{One}}, {{Many}}},
  author = {Argyle, Lisa P. and Busby, Ethan C. and Fulda, Nancy and Gubler, Joshua R. and Rytting, Christopher and Wingate, David},
  date = {2023-07},
  journaltitle = {Political Analysis},
  volume = {31},
  number = {3},
  pages = {337--351},
  issn = {1047-1987, 1476-4989},
  doi = {10.1017/pan.2023.2},
  language = {en}
}

@article{Arseniev-KoehlerTheoreticalFoundationsLimits2024,
  title = {Theoretical {{Foundations}} and {{Limits}} of {{Word Embeddings}}: {{What Types}} of {{Meaning}} Can {{They Capture}}?},
  shorttitle = {Theoretical {{Foundations}} and {{Limits}} of {{Word Embeddings}}},
  author = {Arseniev-Koehler, Alina},
  date = {2024-11-01},
  journaltitle = {Sociological Methods \& Research},
  volume = {53},
  number = {4},
  pages = {1753--1793},
  publisher = {SAGE Publications Inc},
  issn = {0049-1241},
  doi = {10.1177/00491241221140142}
}

@online{AtariWhichHumans2023,
  title = {Which {{Humans}}?},
  author = {Atari, Mohammad and Xue, Mona and Park, Peter and Blasi, Damián and Henrich, Joseph},
  date = {2023-09-22},
  eprinttype = {OSF},
  doi = {10.31234/osf.io/5b26t},
  language = {en-us},
  pubstate = {prepublished}
}

@article{BailCanGenerativeAI2024,
  title = {Can {{Generative AI}} Improve Social Science?},
  author = {Bail, Christopher A.},
  date = {2024-05-21},
  journaltitle = {Proceedings of the National Academy of Sciences},
  volume = {121},
  number = {21},
  pages = {e2314021121},
  publisher = {Proceedings of the National Academy of Sciences},
  doi = {10.1073/pnas.2314021121}
}

@article{BankerMachineassistedSocialPsychology2024,
  title = {Machine-Assisted Social Psychology Hypothesis Generation},
  author = {Banker, Sachin and Chatterjee, Promothesh and Mishra, Himanshu and Mishra, Arul},
  date = {2024-09},
  journaltitle = {American Psychologist},
  shortjournal = {American Psychologist},
  volume = {79},
  number = {6},
  pages = {789--797},
  publisher = {American Psychological Association},
  issn = {0003-066X},
  doi = {10.1037/amp0001222}
}

@article{BisbeeSyntheticReplacementsHuman2024,
  title = {Synthetic {{Replacements}} for {{Human Survey Data}}? {{The Perils}} of {{Large Language Models}}},
  shorttitle = {Synthetic {{Replacements}} for {{Human Survey Data}}?},
  author = {Bisbee, James and Clinton, Joshua D. and Dorff, Cassy and Kenkel, Brenton and Larson, Jennifer M.},
  date = {2024-05-17},
  journaltitle = {Political Analysis},
  pages = {1--16},
  issn = {1047-1987, 1476-4989},
  doi = {10.1017/pan.2024.5},
  language = {en}
}

@article{BorgattiNetworkAnalysisSocial2009,
  title = {Network {{Analysis}} in the {{Social Sciences}}},
  author = {Borgatti, Stephen P. and Mehra, Ajay and Brass, Daniel J. and Labianca, Giuseppe},
  date = {2009-02-13},
  journaltitle = {Science},
  shortjournal = {Science},
  volume = {323},
  number = {5916},
  eprint = {19213908},
  eprinttype = {pubmed},
  pages = {892--895},
  issn = {0036-8075, 1095-9203},
  doi = {10.1126/science.1165821},
  language = {en}
}

@article{BoutylineMeaningHyperspaceWord2025,
  title = {Meaning in {{Hyperspace}}: {{Word Embeddings}} as {{Tools}} for {{Cultural Measurement}}},
  shorttitle = {Meaning in {{Hyperspace}}},
  author = {Boutyline, Andrei and Arseniev-Koehler, Alina},
  date = {2025-07-30},
  journaltitle = {Annual Review of Sociology},
  volume = {51},
  pages = {89--107},
  publisher = {Annual Reviews},
  issn = {0360-0572, 1545-2115},
  doi = {10.1146/annurev-soc-090324-024027},
  issue = {Volume 51, 2025},
  language = {en}
}

@article{BroskaMixedSubjectsDesign2025,
  title = {The {{Mixed Subjects Design}}: {{Treating Large Language Models}} as {{Potentially Informative Observations}}},
  shorttitle = {The {{Mixed Subjects Design}}},
  author = {Broska, David and Howes, Michael and family=Loon, given=Austin, prefix=van, useprefix=true},
  date = {2025-08-01},
  journaltitle = {Sociological Methods \& Research},
  volume = {54},
  number = {3},
  pages = {1074--1109},
  publisher = {SAGE Publications Inc},
  issn = {0049-1241},
  doi = {10.1177/00491241251326865}
}

@online{BrownLanguageModelsAre2020,
  title = {Language {{Models}} Are {{Few-Shot Learners}}},
  author = {Brown, Tom B. and Mann, Benjamin and Ryder, Nick and Subbiah, Melanie and Kaplan, Jared and Dhariwal, Prafulla and Neelakantan, Arvind and Shyam, Pranav and Sastry, Girish and Askell, Amanda and Agarwal, Sandhini and Herbert-Voss, Ariel and Krueger, Gretchen and Henighan, Tom and Child, Rewon and Ramesh, Aditya and Ziegler, Daniel M. and Wu, Jeffrey and Winter, Clemens and Hesse, Christopher and Chen, Mark and Sigler, Eric and Litwin, Mateusz and Gray, Scott and Chess, Benjamin and Clark, Jack and Berner, Christopher and McCandlish, Sam and Radford, Alec and Sutskever, Ilya and Amodei, Dario},
  date = {2020-07-22},
  eprint = {2005.14165},
  eprinttype = {arXiv},
  eprintclass = {cs},
  doi = {10.48550/arXiv.2005.14165},
  pubstate = {prepublished},
  version = {4}
}

@article{CarlsenComputationalGroundedTheory2022,
  title = {Computational Grounded Theory Revisited: {{From}} Computer-Led to Computer-Assisted Text Analysis},
  shorttitle = {Computational Grounded Theory Revisited},
  author = {Carlsen, Hjalmar Bang and Ralund, Snorre},
  date = {2022-01-01},
  journaltitle = {Big Data \& Society},
  volume = {9},
  number = {1},
  pages = {20539517221080146},
  publisher = {SAGE Publications Ltd},
  issn = {2053-9517},
  doi = {10.1177/20539517221080146},
  language = {EN}
}

@online{Chang12BestPractices2024,
  type = {Working Paper},
  title = {12 {{Best Practices}} for {{Leveraging Generative AI}} in {{Experimental Research}}},
  author = {Chang, Samuel and Kennedy, Andrew and Leonard, Aaron and List, John A.},
  date = {2024-10},
  series = {Working {{Paper Series}}},
  number = {33025},
  eprint = {33025},
  eprinttype = {National Bureau of Economic Research},
  doi = {10.3386/w33025},
  pubstate = {prepublished}
}

@article{ChowdhuryGenderDifferencesGiving2017,
  title = {Gender {{Differences}} in the {{Giving}} and {{Taking Variants}} of the {{Dictator Game}}},
  author = {Chowdhury, Subhasish M. and Jeon, Joo Young and Saha, Bibhas},
  date = {2017},
  journaltitle = {Southern Economic Journal},
  volume = {84},
  number = {2},
  pages = {474--483},
  issn = {2325-8012},
  doi = {10.1002/soej.12223},
  language = {en}
}

@article{ChowdhuryGenderDifferencesGiving2020,
  title = {Gender Differences in Giving and the Anticipation Regarding Giving in Dictator Games},
  author = {Chowdhury, Subhasish M and Grossman, Philip J and Jeon, Joo Young},
  date = {2020-07-01},
  journaltitle = {Oxford Economic Papers},
  shortjournal = {Oxford Economic Papers},
  volume = {72},
  number = {3},
  pages = {772--779},
  issn = {0030-7653},
  doi = {10.1093/oep/gpaa002}
}

@online{DaiDeepSeekMoEUltimateExpert2024,
  title = {{{DeepSeekMoE}}: {{Towards Ultimate Expert Specialization}} in {{Mixture-of-Experts Language Models}}},
  shorttitle = {{{DeepSeekMoE}}},
  author = {Dai, Damai and Deng, Chengqi and Zhao, Chenggang and Xu, R. X. and Gao, Huazuo and Chen, Deli and Li, Jiashi and Zeng, Wangding and Yu, Xingkai and Wu, Y. and Xie, Zhenda and Li, Y. K. and Huang, Panpan and Luo, Fuli and Ruan, Chong and Sui, Zhifang and Liang, Wenfeng},
  date = {2024-01-11},
  eprint = {2401.06066},
  eprinttype = {arXiv},
  eprintclass = {cs},
  doi = {10.48550/arXiv.2401.06066},
  pubstate = {prepublished}
}

@online{DeepSeek-AIDeepSeekR1IncentivizingReasoning2025,
  title = {{{DeepSeek-R1}}: {{Incentivizing Reasoning Capability}} in {{LLMs}} via {{Reinforcement Learning}}},
  shorttitle = {{{DeepSeek-R1}}},
  author = {DeepSeek-AI and Guo, Daya and Yang, Dejian and Zhang, Haowei and Song, Junxiao and Zhang, Ruoyu and Xu, Runxin and Zhu, Qihao and Ma, Shirong and Wang, Peiyi and Bi, Xiao and Zhang, Xiaokang and Yu, Xingkai and Wu, Yu and Wu, Z. F. and Gou, Zhibin and Shao, Zhihong and Li, Zhuoshu and Gao, Ziyi and Liu, Aixin and Xue, Bing and Wang, Bingxuan and Wu, Bochao and Feng, Bei and Lu, Chengda and Zhao, Chenggang and Deng, Chengqi and Zhang, Chenyu and Ruan, Chong and Dai, Damai and Chen, Deli and Ji, Dongjie and Li, Erhang and Lin, Fangyun and Dai, Fucong and Luo, Fuli and Hao, Guangbo and Chen, Guanting and Li, Guowei and Zhang, H. and Bao, Han and Xu, Hanwei and Wang, Haocheng and Ding, Honghui and Xin, Huajian and Gao, Huazuo and Qu, Hui and Li, Hui and Guo, Jianzhong and Li, Jiashi and Wang, Jiawei and Chen, Jingchang and Yuan, Jingyang and Qiu, Junjie and Li, Junlong and Cai, J. L. and Ni, Jiaqi and Liang, Jian and Chen, Jin and Dong, Kai and Hu, Kai and Gao, Kaige and Guan, Kang and Huang, Kexin and Yu, Kuai and Wang, Lean and Zhang, Lecong and Zhao, Liang and Wang, Litong and Zhang, Liyue and Xu, Lei and Xia, Leyi and Zhang, Mingchuan and Zhang, Minghua and Tang, Minghui and Li, Meng and Wang, Miaojun and Li, Mingming and Tian, Ning and Huang, Panpan and Zhang, Peng and Wang, Qiancheng and Chen, Qinyu and Du, Qiushi and Ge, Ruiqi and Zhang, Ruisong and Pan, Ruizhe and Wang, Runji and Chen, R. J. and Jin, R. L. and Chen, Ruyi and Lu, Shanghao and Zhou, Shangyan and Chen, Shanhuang and Ye, Shengfeng and Wang, Shiyu and Yu, Shuiping and Zhou, Shunfeng and Pan, Shuting and Li, S. S. and Zhou, Shuang and Wu, Shaoqing and Ye, Shengfeng and Yun, Tao and Pei, Tian and Sun, Tianyu and Wang, T. and Zeng, Wangding and Zhao, Wanjia and Liu, Wen and Liang, Wenfeng and Gao, Wenjun and Yu, Wenqin and Zhang, Wentao and Xiao, W. L. and An, Wei and Liu, Xiaodong and Wang, Xiaohan and Chen, Xiaokang and Nie, Xiaotao and Cheng, Xin and Liu, Xin and Xie, Xin and Liu, Xingchao and Yang, Xinyu and Li, Xinyuan and Su, Xuecheng and Lin, Xuheng and Li, X. Q. and Jin, Xiangyue and Shen, Xiaojin and Chen, Xiaosha and Sun, Xiaowen and Wang, Xiaoxiang and Song, Xinnan and Zhou, Xinyi and Wang, Xianzu and Shan, Xinxia and Li, Y. K. and Wang, Y. Q. and Wei, Y. X. and Zhang, Yang and Xu, Yanhong and Li, Yao and Zhao, Yao and Sun, Yaofeng and Wang, Yaohui and Yu, Yi and Zhang, Yichao and Shi, Yifan and Xiong, Yiliang and He, Ying and Piao, Yishi and Wang, Yisong and Tan, Yixuan and Ma, Yiyang and Liu, Yiyuan and Guo, Yongqiang and Ou, Yuan and Wang, Yuduan and Gong, Yue and Zou, Yuheng and He, Yujia and Xiong, Yunfan and Luo, Yuxiang and You, Yuxiang and Liu, Yuxuan and Zhou, Yuyang and Zhu, Y. X. and Xu, Yanhong and Huang, Yanping and Li, Yaohui and Zheng, Yi and Zhu, Yuchen and Ma, Yunxian and Tang, Ying and Zha, Yukun and Yan, Yuting and Ren, Z. Z. and Ren, Zehui and Sha, Zhangli and Fu, Zhe and Xu, Zhean and Xie, Zhenda and Zhang, Zhengyan and Hao, Zhewen and Ma, Zhicheng and Yan, Zhigang and Wu, Zhiyu and Gu, Zihui and Zhu, Zijia and Liu, Zijun and Li, Zilin and Xie, Ziwei and Song, Ziyang and Pan, Zizheng and Huang, Zhen and Xu, Zhipeng and Zhang, Zhongyu and Zhang, Zhen},
  date = {2025-01-22},
  eprint = {2501.12948},
  eprinttype = {arXiv},
  eprintclass = {cs},
  doi = {10.48550/arXiv.2501.12948},
  pubstate = {prepublished}
}

@article{Donate-BuendiaGenderOtherModerators2022,
  title = {Gender and Other Moderators of Giving in the Dictator Game: {{A}} Meta-Analysis},
  shorttitle = {Gender and Other Moderators of Giving in the Dictator Game},
  author = {Doñate-Buendía, Anabel and García-Gallego, Aurora and Petrović, Marko},
  date = {2022-06-01},
  journaltitle = {Journal of Economic Behavior \& Organization},
  shortjournal = {Journal of Economic Behavior \& Organization},
  volume = {198},
  pages = {280--301},
  issn = {0167-2681},
  doi = {10.1016/j.jebo.2022.03.031}
}

@article{EaglyHisHersProsocial2009,
  title = {The His and Hers of Prosocial Behavior: {{An}} Examination of the Social Psychology of Gender},
  shorttitle = {The His and Hers of Prosocial Behavior},
  author = {Eagly, Alice H.},
  date = {2009-11},
  journaltitle = {American Psychologist},
  shortjournal = {American Psychologist},
  volume = {64},
  number = {8},
  pages = {644--658},
  publisher = {American Psychological Association},
  issn = {0003-066X},
  doi = {10.1037/0003-066X.64.8.644}
}

@article{EdelmannComputationalSocialScience2020,
  title = {Computational {{Social Science}} and {{Sociology}}},
  author = {Edelmann, Achim and Wolff, Tom and Montagne, Danielle and Bail, Christopher A.},
  date = {2020-07-30},
  journaltitle = {Annual Review of Sociology},
  shortjournal = {Annu. Rev. Sociol.},
  volume = {46},
  number = {1},
  pages = {61--81},
  publisher = {Annual Reviews},
  issn = {0360-0572},
  doi = {10.1146/annurev-soc-121919-054621}
}

@online{EngelsNotAllLanguage2024,
  title = {Not {{All Language Model Features Are Linear}}},
  author = {Engels, Joshua and Michaud, Eric J. and Liao, Isaac and Gurnee, Wes and Tegmark, Max},
  date = {2024-10-08},
  eprint = {2405.14860},
  eprinttype = {arXiv},
  doi = {10.48550/arXiv.2405.14860},
  pubstate = {prepublished}
}

@book{FrickerEpistemicInjusticePower2007,
  title = {Epistemic {{Injustice}}: {{Power}} and the {{Ethics}} of {{Knowing}}},
  shorttitle = {Epistemic {{Injustice}}},
  author = {Fricker, Miranda},
  date = {2007-07-05},
  eprint = {lncSDAAAQBAJ},
  eprinttype = {googlebooks},
  publisher = {Clarendon Press},
  isbn = {978-0-19-823790-7},
  language = {en},
  pagetotal = {199}
}

@article{GargWordEmbeddingsQuantify2018,
  title = {Word Embeddings Quantify 100 Years of Gender and Ethnic Stereotypes},
  author = {Garg, Nikhil and Schiebinger, Londa and Jurafsky, Dan and Zou, James},
  date = {2018-04-17},
  journaltitle = {Proceedings of the National Academy of Sciences},
  shortjournal = {PNAS},
  volume = {115},
  number = {16},
  pages = {E3635-E3644},
  issn = {0027-8424, 1091-6490},
  doi = {10.1073/pnas.1720347115},
  language = {en}
}

@article{GrimmerTextDataPromise2013,
  title = {Text as {{Data}}: {{The Promise}} and {{Pitfalls}} of {{Automatic Content Analysis Methods}} for {{Political Texts}}},
  shorttitle = {Text as {{Data}}},
  author = {Grimmer, Justin and Stewart, Brandon M.},
  date = {2013},
  journaltitle = {Political Analysis},
  volume = {21},
  number = {3},
  pages = {267--297},
  issn = {1047-1987, 1476-4989},
  doi = {10.1093/pan/mps028},
  language = {en}
}

@online{GurneeLanguageModelsRepresent2024,
  title = {Language {{Models Represent Space}} and {{Time}}},
  author = {Gurnee, Wes and Tegmark, Max},
  date = {2024-03-04},
  eprint = {2310.02207},
  eprinttype = {arXiv},
  doi = {10.48550/arXiv.2310.02207},
  pubstate = {prepublished}
}

@online{HortonLargeLanguageModels2023,
  type = {Working Paper},
  title = {Large {{Language Models}} as {{Simulated Economic Agents}}: {{What Can We Learn}} from {{Homo Silicus}}?},
  shorttitle = {Large {{Language Models}} as {{Simulated Economic Agents}}},
  author = {Horton, John J.},
  date = {2023-04},
  series = {Working {{Paper Series}}},
  number = {31122},
  eprint = {31122},
  eprinttype = {National Bureau of Economic Research},
  doi = {10.3386/w31122},
  pubstate = {prepublished}
}

@online{JohnsonEvidenceBehaviorConsistent2023,
  title = {Evidence of Behavior Consistent with Self-Interest and Altruism in an Artificially Intelligent Agent},
  author = {Johnson, Tim and Obradovich, Nick},
  date = {2023-01-05},
  eprint = {2301.02330},
  eprinttype = {arXiv},
  doi = {10.48550/arXiv.2301.02330},
  pubstate = {prepublished}
}

@inproceedings{KeaheyLessonsLearnedChameleon2020,
  title = {Lessons {{Learned}} from the {{Chameleon Testbed}}},
  author = {Keahey, Kate and Anderson, Jason and Zhen, Zhuo and Riteau, Pierre and Ruth, Paul and Stanzione, Dan and Cevik, Mert and Colleran, Jacob and Gunawi, Haryadi S. and Hammock, Cody and Mambretti, Joe and Barnes, Alexander and Halbah, François and Rocha, Alex and Stubbs, Joe},
  date = {2020},
  pages = {219--233},
  url = {https://www.usenix.org/conference/atc20/presentation/keahey},
  urldate = {2021-09-05},
  eventtitle = {2020 \{\vphantom\}{{USENIX}}\vphantom\{\} {{Annual Technical Conference}} (\{\vphantom\}{{USENIX}}\vphantom\{\} \{\vphantom\}{{ATC}}\vphantom\{\} 20)},
  isbn = {978-1-939133-14-4},
  language = {en}
}

@online{KimLinearRepresentationsPolitical2025,
  title = {Linear {{Representations}} of {{Political Perspective Emerge}} in {{Large Language Models}}},
  author = {Kim, Junsol and Evans, James and Schein, Aaron},
  date = {2025-04-02},
  eprint = {2503.02080},
  eprinttype = {arXiv},
  eprintclass = {cs},
  doi = {10.48550/arXiv.2503.02080},
  pubstate = {prepublished}
}

@report{KorinekAIAgentsEconomic2025,
  title = {{{AI Agents}} for {{Economic Research}}},
  author = {Korinek, Anton},
  date = {2025-09-08},
  number = {w34202},
  institution = {National Bureau of Economic Research},
  doi = {10.3386/w34202},
  language = {en}
}

@article{KozlowskiGeometryCultureAnalyzing2019,
  title = {The {{Geometry}} of {{Culture}}: {{Analyzing}} the {{Meanings}} of {{Class}} through {{Word Embeddings}}},
  shorttitle = {The {{Geometry}} of {{Culture}}},
  author = {Kozlowski, Austin C. and Taddy, Matt and Evans, James A.},
  date = {2019-10-01},
  journaltitle = {American Sociological Review},
  shortjournal = {Am Sociol Rev},
  volume = {84},
  number = {5},
  pages = {905--949},
  issn = {0003-1224},
  doi = {10.1177/0003122419877135},
  language = {en}
}

@article{KozlowskiSimulatingSubjectsPromise2025,
  title = {Simulating {{Subjects}}: {{The Promise}} and {{Peril}} of {{Artificial Intelligence Stand-Ins}} for {{Social Agents}} and {{Interactions}}},
  shorttitle = {Simulating {{Subjects}}},
  author = {Kozlowski, Austin C. and Evans, James},
  date = {2025-08-01},
  journaltitle = {Sociological Methods \& Research},
  volume = {54},
  number = {3},
  pages = {1017--1073},
  publisher = {SAGE Publications Inc},
  issn = {0049-1241},
  doi = {10.1177/00491241251337316},
  language = {EN}
}

@article{LakeBuildingMachinesThat2017,
  title = {Building Machines That Learn and Think like People},
  author = {Lake, Brenden M. and Ullman, Tomer D. and Tenenbaum, Joshua B. and Gershman, Samuel J.},
  date = {2017-01},
  journaltitle = {Behavioral and Brain Sciences},
  volume = {40},
  pages = {e253},
  issn = {0140-525X, 1469-1825},
  doi = {10.1017/S0140525X16001837},
  language = {en}
}

@online{LengLLMAgentsExhibit2024,
  title = {Do {{LLM Agents Exhibit Social Behavior}}?},
  author = {Leng, Yan and Yuan, Yuan},
  date = {2024-10-15},
  eprint = {2312.15198},
  eprinttype = {arXiv},
  eprintclass = {cs},
  doi = {10.48550/arXiv.2312.15198},
  pubstate = {prepublished}
}

@online{LepikhinGShardScalingGiant2020,
  title = {{{GShard}}: {{Scaling Giant Models}} with {{Conditional Computation}} and {{Automatic Sharding}}},
  shorttitle = {{{GShard}}},
  author = {Lepikhin, Dmitry and Lee, HyoukJoong and Xu, Yuanzhong and Chen, Dehao and Firat, Orhan and Huang, Yanping and Krikun, Maxim and Shazeer, Noam and Chen, Zhifeng},
  date = {2020-06-30},
  eprint = {2006.16668},
  eprinttype = {arXiv},
  eprintclass = {cs},
  doi = {10.48550/arXiv.2006.16668},
  pubstate = {prepublished}
}

@online{MaCanMachinesThink2024,
  title = {Can {{Machines Think Like Humans}}? {{A Behavioral Evaluation}} of {{LLM-Agents}} in {{Dictator Games}}},
  shorttitle = {Can {{Machines Think Like Humans}}?},
  author = {Ma, Ji},
  date = {2024-12-16},
  eprint = {2410.21359},
  eprinttype = {arXiv},
  eprintclass = {cs},
  doi = {10.48550/arXiv.2410.21359},
  pubstate = {prepublished}
}

@article{MeiTuringTestWhether2024,
  title = {A {{Turing}} Test of Whether {{AI}} Chatbots Are Behaviorally Similar to Humans},
  author = {Mei, Qiaozhu and Xie, Yutong and Yuan, Walter and Jackson, Matthew O.},
  date = {2024-02-27},
  journaltitle = {Proceedings of the National Academy of Sciences},
  volume = {121},
  number = {9},
  pages = {e2313925121},
  publisher = {Proceedings of the National Academy of Sciences},
  doi = {10.1073/pnas.2313925121}
}

@unpublished{MikolovEfficientEstimationWord2013,
  ids = {MikolovEfficientEstimationWord2013a},
  title = {Efficient {{Estimation}} of {{Word Representations}} in {{Vector Space}}},
  author = {Mikolov, Tomas and Chen, Kai and Corrado, Greg and Dean, Jeffrey},
  date = {2013-01-16},
  eprint = {1301.3781},
  eprinttype = {arXiv},
  eprintclass = {cs},
  url = {http://arxiv.org/abs/1301.3781},
  urldate = {2019-05-08}
}

@article{MostafaviContextualEmbeddingsSociological2025,
  title = {Contextual {{Embeddings}} in {{Sociological Research}}: {{Expanding}} the {{Analysis}} of {{Sentiment}} and {{Social Dynamics}}},
  shorttitle = {Contextual {{Embeddings}} in {{Sociological Research}}},
  author = {Mostafavi, Moeen and Porter, Michael D. and Robinson, Dawn T.},
  date = {2025-02-01},
  journaltitle = {Sociological Methodology},
  volume = {55},
  number = {1},
  pages = {25--58},
  publisher = {SAGE Publications Inc},
  issn = {0081-1750},
  doi = {10.1177/00811750241260729}
}

@article{NelsonComputationalGroundedTheory2020,
  title = {Computational {{Grounded Theory}}: {{A Methodological Framework}}},
  shorttitle = {Computational {{Grounded Theory}}},
  author = {Nelson, Laura K.},
  date = {2020-02-01},
  journaltitle = {Sociological Methods \& Research},
  volume = {49},
  number = {1},
  pages = {3--42},
  publisher = {SAGE Publications Inc},
  issn = {0049-1241},
  doi = {10.1177/0049124117729703},
  language = {en}
}

@online{ParkLinearRepresentationHypothesis2024,
  title = {The {{Linear Representation Hypothesis}} and the {{Geometry}} of {{Large Language Models}}},
  author = {Park, Kiho and Choe, Yo Joong and Veitch, Victor},
  date = {2024-07-17},
  eprint = {2311.03658},
  eprinttype = {arXiv},
  doi = {10.48550/arXiv.2311.03658},
  pubstate = {prepublished}
}

@article{RisterPortinariMarancaCorrectingMeasurementErrors2025,
  title = {Correcting the {{Measurement Errors}} of {{AI-Assisted Labeling}} in {{Image Analysis Using Design-Based Supervised Learning}}},
  author = {Rister Portinari Maranca, Alessandra and Chung, Jihoon and Hinck, Musashi and Wolsky, Adam D. and Egami, Naoki and Stewart, Brandon M.},
  date = {2025-08-01},
  journaltitle = {Sociological Methods \& Research},
  volume = {54},
  number = {3},
  pages = {984--1016},
  publisher = {SAGE Publications Inc},
  issn = {0049-1241},
  doi = {10.1177/00491241251333372}
}

@article{RobertsIntroductionVirtualIssue2016,
  title = {Introduction to the {{Virtual Issue}}: {{Recent Innovations}} in {{Text Analysis}} for {{Social Science}}},
  shorttitle = {Introduction to the {{Virtual Issue}}},
  author = {Roberts, Margaret E.},
  date = {2016-04},
  journaltitle = {Political Analysis},
  volume = {24},
  number = {V10},
  pages = {1--5},
  issn = {1047-1987, 1476-4989},
  doi = {10.1017/S1047198700014418},
  language = {en}
}

@online{SennrichNeuralMachineTranslation2016,
  title = {Neural {{Machine Translation}} of {{Rare Words}} with {{Subword Units}}},
  author = {Sennrich, Rico and Haddow, Barry and Birch, Alexandra},
  date = {2016-06-10},
  eprint = {1508.07909},
  eprinttype = {arXiv},
  eprintclass = {cs},
  doi = {10.48550/arXiv.1508.07909},
  pubstate = {prepublished}
}

@inproceedings{SteckCosineSimilarityEmbeddingsReally2024,
  title = {Is {{Cosine-Similarity}} of {{Embeddings Really About Similarity}}?},
  booktitle = {Companion {{Proceedings}} of the {{ACM Web Conference}} 2024},
  author = {Steck, Harald and Ekanadham, Chaitanya and Kallus, Nathan},
  date = {2024-05-13},
  series = {{{WWW}} '24},
  pages = {887--890},
  publisher = {Association for Computing Machinery},
  location = {New York, NY, USA},
  doi = {10.1145/3589335.3651526},
  isbn = {979-8-4007-0172-6}
}

@book{StoltzMappingTextsComputational2024,
  title = {Mapping {{Texts}}: {{Computational Text Analysis}} for the {{Social Sciences}}},
  shorttitle = {Mapping {{Texts}}},
  author = {Stoltz, Dustin S. and Taylor, Marshall A.},
  date = {2024-02-15},
  series = {Computational {{Social Science}}},
  publisher = {Oxford University Press},
  location = {Oxford, New York},
  isbn = {978-0-19-775688-1},
  pagetotal = {328}
}

@online{SubramaniExtractingLatentSteering2022,
  title = {Extracting {{Latent Steering Vectors}} from {{Pretrained Language Models}}},
  author = {Subramani, Nishant and Suresh, Nivedita and Peters, Matthew E.},
  date = {2022-05-10},
  eprint = {2205.05124},
  eprinttype = {arXiv},
  eprintclass = {cs},
  doi = {10.48550/arXiv.2205.05124},
  pubstate = {prepublished}
}

@online{TiggesLinearRepresentationsSentiment2023,
  title = {Linear {{Representations}} of {{Sentiment}} in {{Large Language Models}}},
  author = {Tigges, Curt and Hollinsworth, Oskar John and Geiger, Atticus and Nanda, Neel},
  date = {2023-10-23},
  eprint = {2310.15154},
  eprinttype = {arXiv},
  doi = {10.48550/arXiv.2310.15154},
  pubstate = {prepublished}
}

@online{TurnerSteeringLanguageModels2024,
  title = {Steering {{Language Models With Activation Engineering}}},
  author = {Turner, Alexander Matt and Thiergart, Lisa and Leech, Gavin and Udell, David and Vazquez, Juan J. and Mini, Ulisse and MacDiarmid, Monte},
  date = {2024-10-10},
  eprint = {2308.10248},
  eprinttype = {arXiv},
  doi = {10.48550/arXiv.2308.10248},
  pubstate = {prepublished},
  version = {5}
}

@inproceedings{VaswaniAttentionAllYou2017,
  title = {Attention Is {{All}} You {{Need}}},
  booktitle = {Advances in {{Neural Information Processing Systems}}},
  author = {Vaswani, Ashish and Shazeer, Noam and Parmar, Niki and Uszkoreit, Jakob and Jones, Llion and Gomez, Aidan N and Kaiser, Łukasz and Polosukhin, Illia},
  date = {2017},
  volume = {30},
  publisher = {Curran Associates, Inc.},
  url = {https://proceedings.neurips.cc/paper_files/paper/2017/hash/3f5ee243547dee91fbd053c1c4a845aa-Abstract.html},
  urldate = {2024-03-15}
}

@article{WangLargeLanguageModels2025,
  title = {Large Language Models That Replace Human Participants Can Harmfully Misportray and Flatten Identity Groups},
  author = {Wang, Angelina and Morgenstern, Jamie and Dickerson, John P.},
  date = {2025-03},
  journaltitle = {Nature Machine Intelligence},
  shortjournal = {Nat Mach Intell},
  volume = {7},
  number = {3},
  pages = {400--411},
  publisher = {Nature Publishing Group},
  issn = {2522-5839},
  doi = {10.1038/s42256-025-00986-z},
  language = {en}
}

@inproceedings{XieCanLargeLanguage2024,
  title = {Can {{Large Language Model Agents Simulate Human Trust Behaviors}}?},
  author = {Xie, Chengxing and Chen, Canyu and Jia, Feiran and Ye, Ziyu and Shu, Kai and Bibi, Adel and Hu, Ziniu and Torr, Philip and Ghanem, Bernard and Li, Guohao and Lai, Shiyang and Evans, James},
  date = {2024-03-10},
  eprint = {2402.04559},
  eprinttype = {arXiv},
  publisher = {arXiv},
  location = {Miami, Florida, USA},
  doi = {10.48550/arXiv.2402.04559},
  eventtitle = {Empirical {{Methods}} in {{Natural Language Processing}}}
}

@online{ZhouHypothesisGenerationLarge2024,
  title = {Hypothesis {{Generation}} with {{Large Language Models}}},
  author = {Zhou, Yangqiaoyu and Liu, Haokun and Srivastava, Tejes and Mei, Hongyuan and Tan, Chenhao},
  date = {2024-08-23},
  eprint = {2404.04326},
  eprinttype = {arXiv},
  eprintclass = {cs},
  doi = {10.48550/arXiv.2404.04326},
  pubstate = {prepublished}
}

@article{ZiemsCanLargeLanguage2024,
  title = {Can {{Large Language Models Transform Computational Social Science}}?},
  author = {Ziems, Caleb and Held, William and Shaikh, Omar and Chen, Jiaao and Zhang, Zhehao and Yang, Diyi},
  date = {2024-03-01},
  journaltitle = {Computational Linguistics},
  shortjournal = {Computational Linguistics},
  volume = {50},
  number = {1},
  pages = {237--291},
  issn = {0891-2017},
  doi = {10.1162/coli_a_00502}
}
\endgroup

\clearpage
\begingroup
\section*{\textsc{Online Appendix}}
\begin{appendix}
  \begin{refsection}
    \renewcommand\thetable{\Alph{section}\arabic{table}}
\renewcommand\thefigure{\Alph{section}\arabic{figure}}
\setcounter{footnote}{0}
\setcounter{page}{1}
\setcounter{table}{0}
\setcounter{figure}{0}

\begin{center}
    \centering\singlespacing\textsc{{\mytitle}}\\~\\
\end{center}

\vspace{-2cm}

\begingroup
\onehalfspacing
\normalsize
\etocdepthtag.toc{mtappendix}
\etocsettagdepth{mtchapter}{none}
\etocsettagdepth{mtappendix}{subsubsection}
\setcounter{tocdepth}{3}
\tableofcontents
\endgroup

\clearpage
\section{Results}
\setcounter{table}{0}
\setcounter{figure}{0}

\begin{table}[!htbp]
    \centering
    \caption{\textsc{Distribution of Categorical Variables in Baseline}} \label{tab:baseline_distribution}
    \begin{tabular}{rrr}
        \hline\hline
        {Variable} & {Counts} & {Percentages}           \\ \hline
                   &          &                         \\[-0.5em]
        \multicolumn{1}{l}{\textit{Gender}}             \\
        Male       & 302      & 52.9\%                  \\
        Female     & 269      & 47.1\%                  \\
                   &          &                         \\[-0.5em]
        \multicolumn{1}{l}{\textit{Game Type}}          \\
        Give       & 334      & 58.5\%                  \\
        Take       & 237      & 41.5\%                  \\
                   &          &                         \\[-0.5em]
        \multicolumn{1}{l}{\textit{Social Distance}}    \\
        Stranger   & 329      & 57.6\%                  \\
        Meet       & 242      & 42.4\%                  \\
                   &          &                         \\[-0.5em]
        \multicolumn{1}{l}{\textit{Amount Transferred}} \\
        0          & 200      & 35.0\%                  \\
        10         & 371      & 65.0\%                  \\
                   &          &                         \\[-0.5em]
        \hline\hline
    \end{tabular}
\end{table}

\begin{figure}[!htbp]
    \centering
    \includegraphics[width=.8\textwidth]{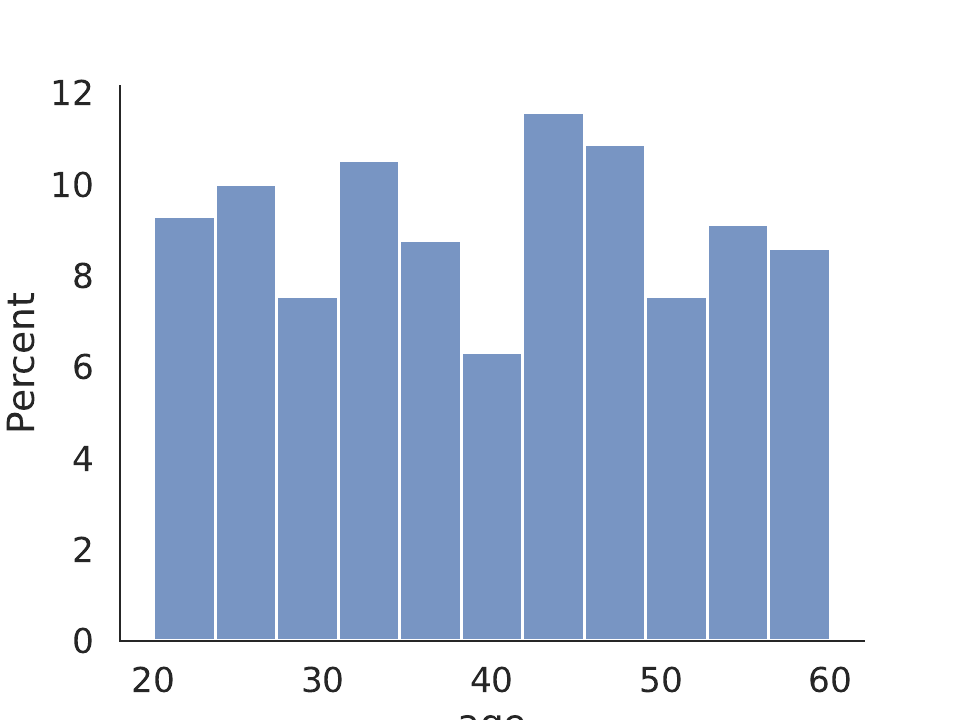}
    \caption{\textsc{Distribution of Age in Baseline}} \label{fig:age_distribution}
\end{figure}

\begin{sidewaysfigure}
    \centering
    \includegraphics[width=.8\textwidth]{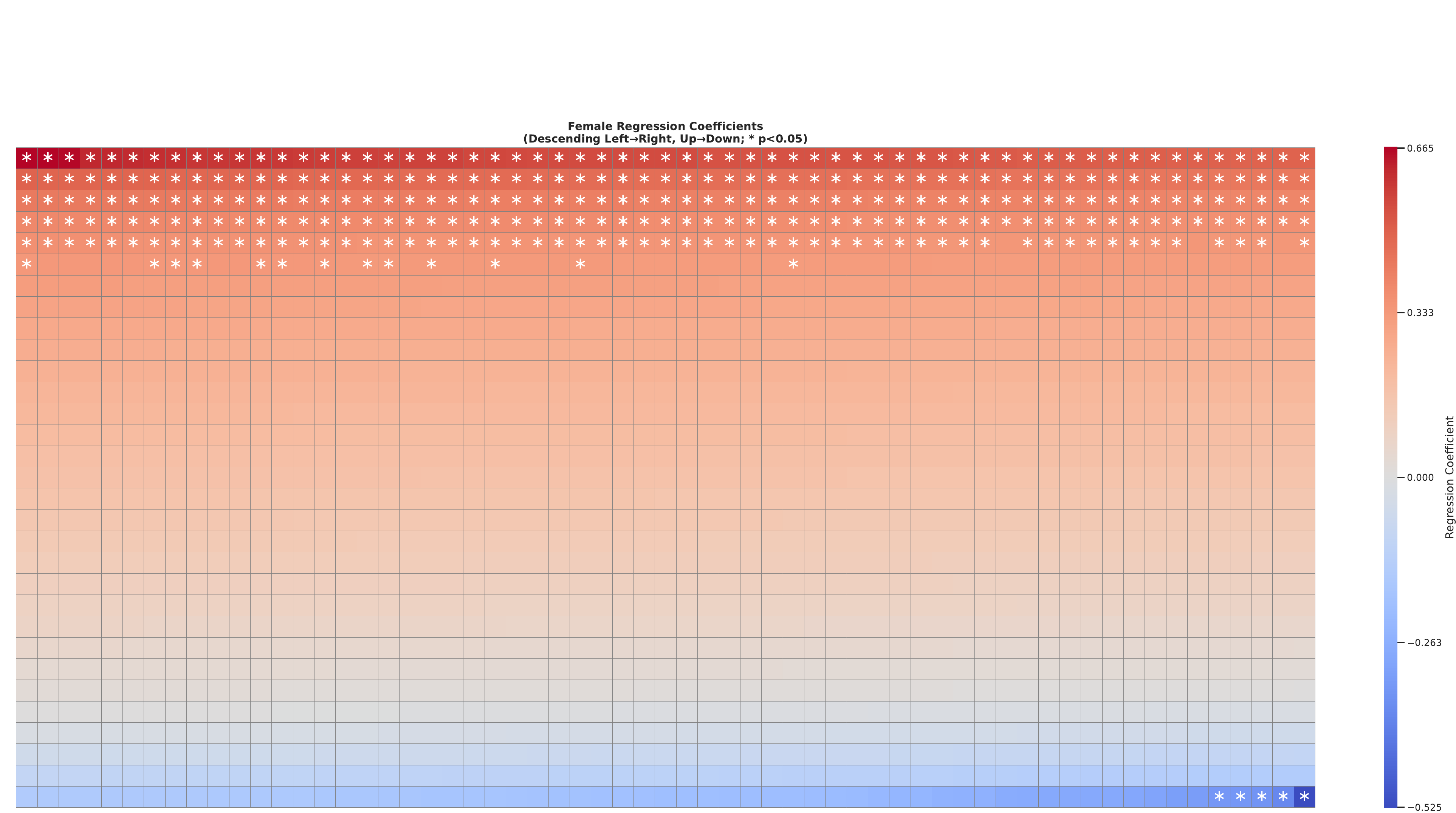}
    \caption{\textsc{Ordered Heatmap of Female Regression Coefficients}} \label{fig:rc_heatmap_female_ordered}
\end{sidewaysfigure}

\clearpage
\section{Results of Orthogonality Analysis}
\setcounter{table}{0}
\setcounter{figure}{0}

\begin{figure}[!htbp]
    \centering
    \includegraphics[width=1\textwidth]{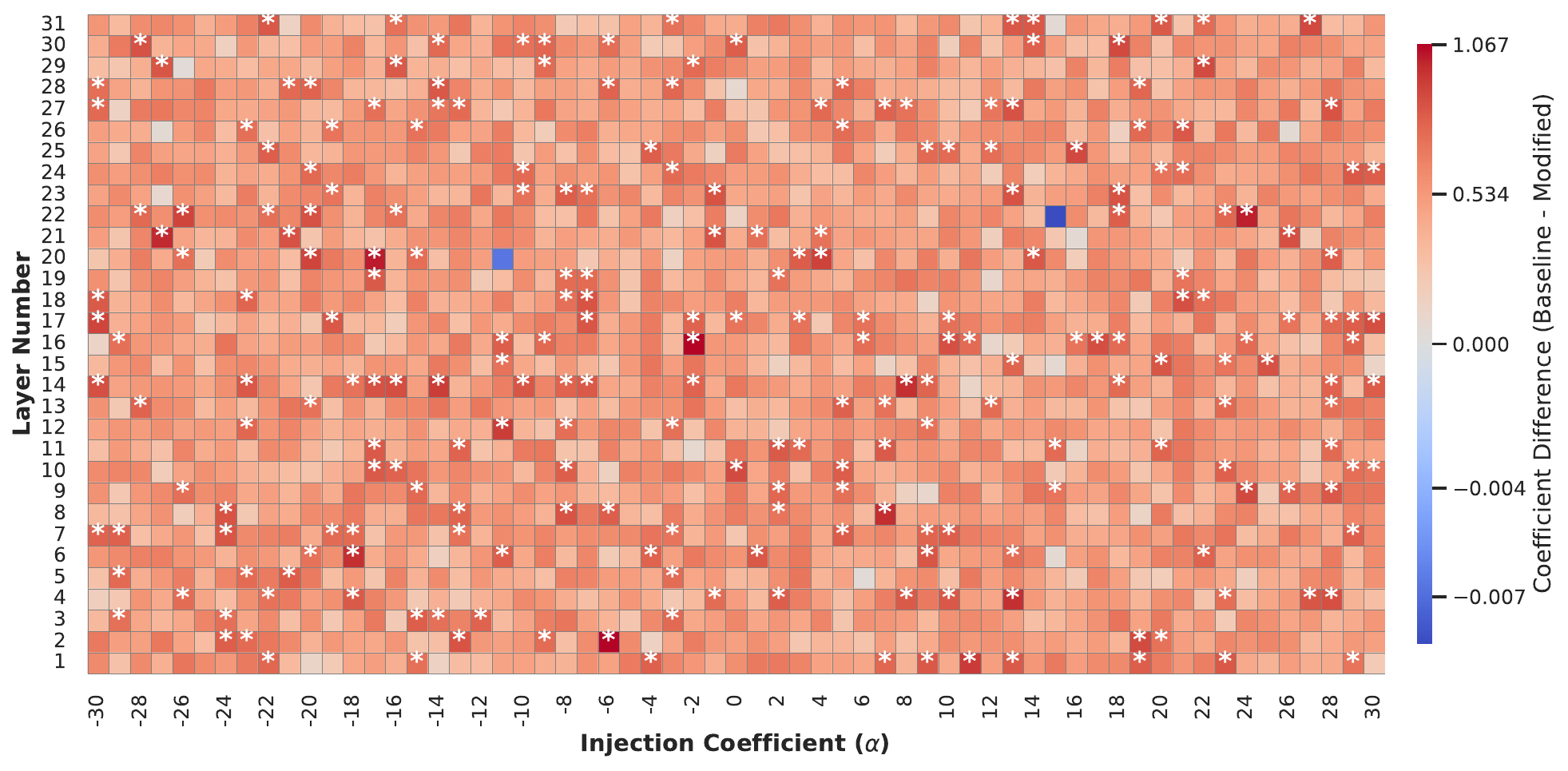}
    \caption{\textsc{Difference of Regression Coefficients: ${Meet-Stranger}$}}
    \label{fig:coef_diff_heatmap_meet}
    \begin{minipage}{0.9\textwidth}
        \footnotesize
        \textit{Notes}: * 95\% CIs not overlapping between baseline and manipulation.
    \end{minipage}
\end{figure}

\begin{figure}[!htbp]
    \centering
    \includegraphics[width=1\textwidth]{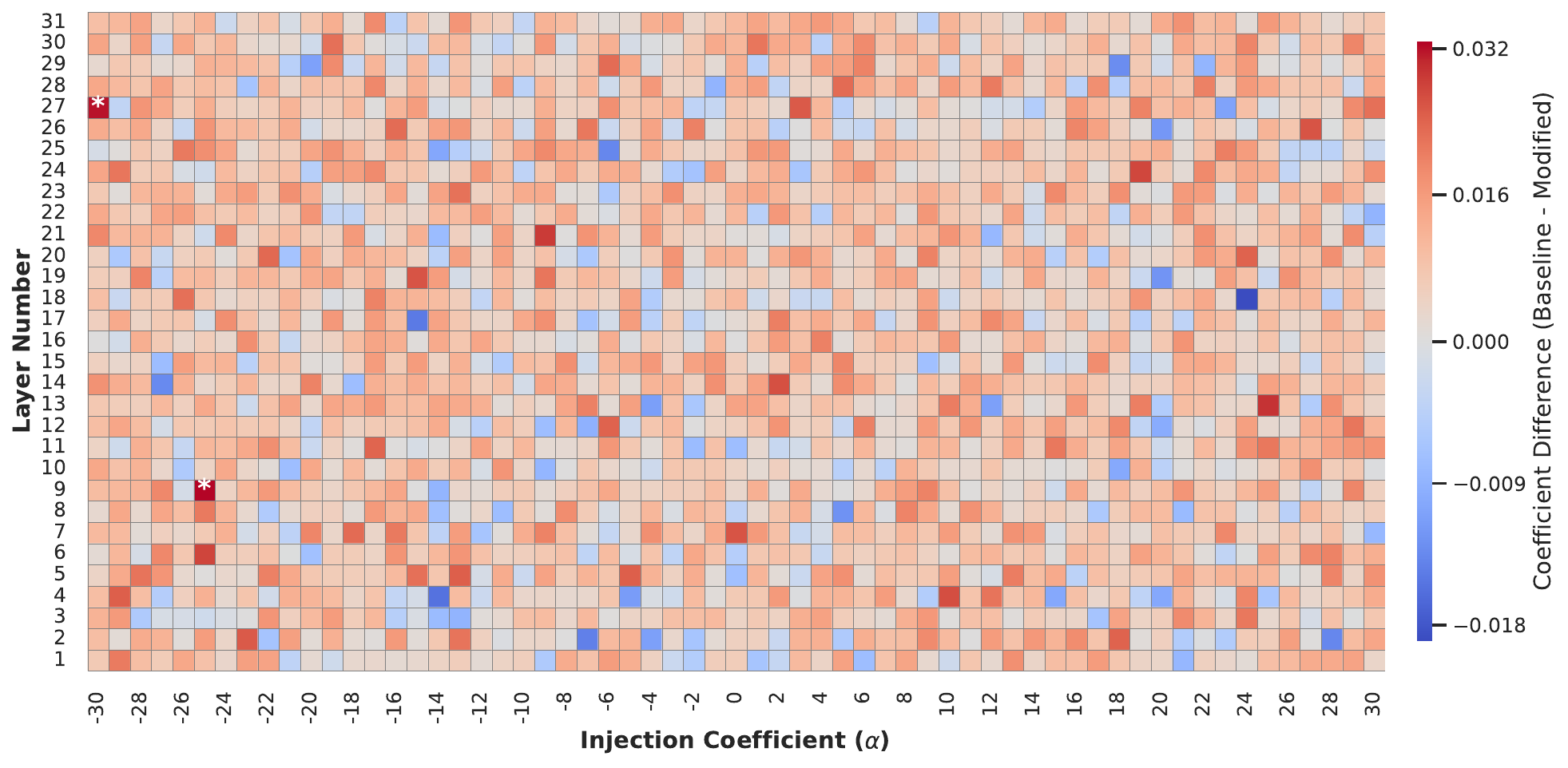}
    \caption{\textsc{Difference of Regression Coefficients: ${Age}$}}
    \label{fig:coef_diff_heatmap_age}
    \begin{minipage}{0.9\textwidth}
        \footnotesize
        \textit{Notes}: * 95\% CIs not overlapping between baseline and manipulation.
    \end{minipage}
\end{figure}

\begin{figure}[!htbp]
    \centering
    \includegraphics[width=1\textwidth]{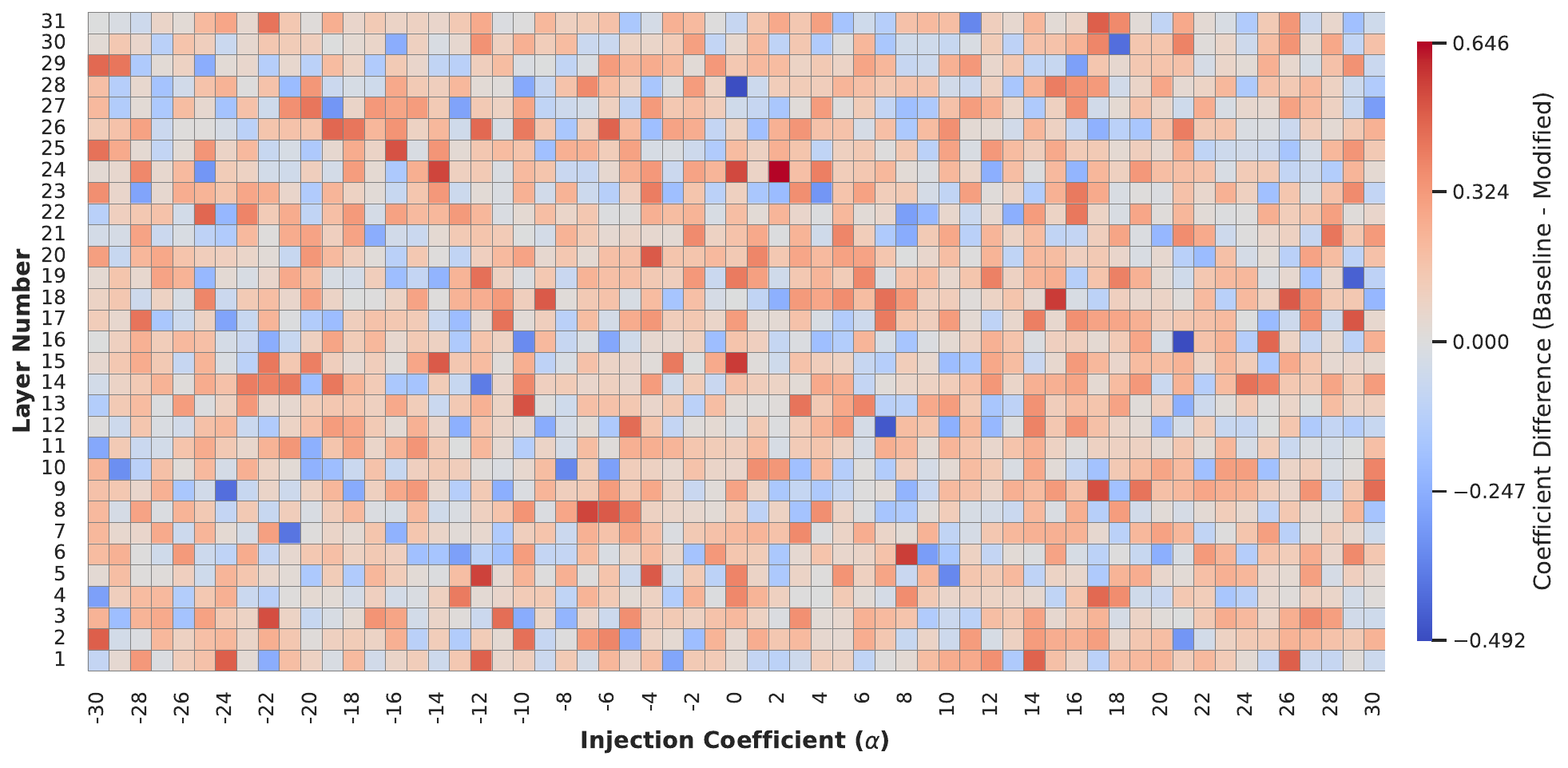}
    \caption{\textsc{Difference of Regression Coefficients: ${Give}$}}
    \label{fig:coef_diff_heatmap_give}
    \begin{minipage}{0.9\textwidth}
        \footnotesize
        \textit{Notes}: * 95\% CIs not overlapping between baseline and manipulation.
    \end{minipage}
\end{figure}

  \end{refsection}
\end{appendix}
\endgroup

\end{document}